\documentclass[11pt,a4paper]{article}
\usepackage[hyperref]{acl2021}
\usepackage{times}
\usepackage{latexsym}
\usepackage{natbib}
\usepackage[multiple]{footmisc}

\usepackage{lipsum}
\newcommand\blfootnote[1]{%
  \begingroup
  \renewcommand\thefootnote{}\footnote{#1}%
  \addtocounter{footnote}{-1}%
  \endgroup
}

\newcommand{\dataname}{\textsf{HarMeme}}
\newcommand{\imgsrc}[2]{\href{#1}{Source #2}}
\newcommand{\imgur}[1]{\href{https://imgur.com/tos}{License #1}}
\newcommand{\ccsnd}[1]{\href{https://creativecommons.org/licenses/by-sa/2.0/deed.en}{License #1}}

\usepackage{multirow}
\usepackage{subcaption}
\usepackage{microtype}
\usepackage{graphicx}
\usepackage{multirow}
\usepackage{enumitem}
\usepackage{rotating}
\usepackage{chngcntr}
\usepackage{amsmath}

\aclfinalcopy % Uncomment this line for the final submission
\title{Detecting Harmful Memes and Their Targets}

\author{Shraman Pramanick$^1$, Dimitar Dimitrov$^2$, Rituparna Mukherjee$^1$, Shivam Sharma$^{1,3}$,\\ \textbf{Md. Shad Akhtar$^1$, Preslav Nakov$^4$, Tanmoy Chakraborty$^1$}\\
  $^1$Indraprastha Institute of Information Technology - Delhi, India  \\
  $^2$Sofia University, Bulgaria \\
  $^3$Wipro AI Labs, India\\
  $^4$Qatar Computing Research Institute, HBKU, Doha, Qatar \\
  \small\texttt{\{shramanp, shivams, shad.akhtar, tanmoy\}@iiitd.ac.in}\\\small\texttt{mitko.bg.ss@gmail.com, ritumukherjee23@gmail.com, pnakov@hbku.edu.qa}}

\date{}

\begin{document}
\maketitle
\begin{abstract}
Among the various modes of communication in social media, the use of Internet memes has emerged as a powerful means to convey political, psychological, and socio-cultural opinions. Although memes are typically humorous in nature, recent days have witnessed a proliferation of {\em harmful memes} targeted to abuse various social entities. As most harmful memes are highly satirical and abstruse without appropriate contexts, off-the-shelf multimodal models may not be adequate to understand their underlying semantics.   
In this work, we propose two novel problem formulations: {\em detecting harmful memes} and {\em the social entities that these harmful memes target}.  
To this end, we present \dataname, the first  benchmark dataset, containing $3,544$ memes related to COVID-19. Each meme went through a rigorous two-stage annotation process. In the first stage, we labeled a meme as {\em very harmful}, {\em partially harmful}, or {\em harmless}; in the second stage, we further annotated the type of target(s) that each harmful meme points to: {\em individual}, {\em organization}, {\em community}, or {\em society/general public/other}. The evaluation results using ten unimodal and multimodal models highlight the importance of using multimodal signals for both tasks. We further discuss the limitations of these models and we argue that more research is needed to address these problems. \blfootnote{WARNING: This paper contains meme examples and words that are offensive in nature.}
\end{abstract}

\begin{figure*}[ht]
\centering
\subfloat[{\tt [0]}\label{fig:meme:exm:1}]{
\includegraphics[width=0.185\textwidth, height=0.185\textwidth]{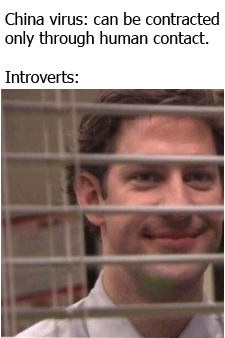}}\hspace{0.2em}
\subfloat[{\centering \tt [2,0]}\label{fig:meme:exm:2}]{
\includegraphics[width=0.185\textwidth, height=0.185\textwidth]{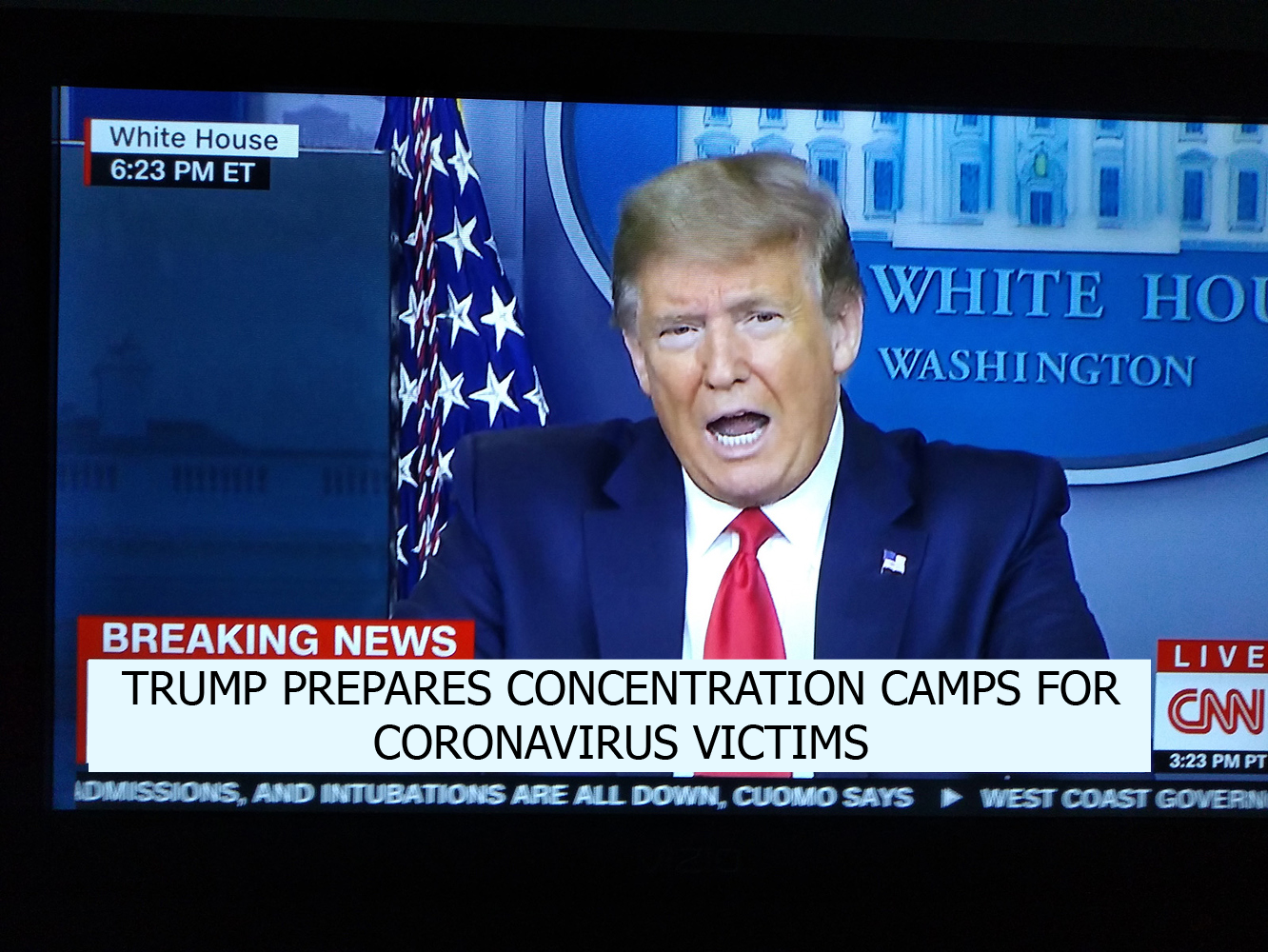}}\hspace{0.2em}
\subfloat[{\tt [1,1]}\label{fig:meme:exm:3}]{
\includegraphics[width=0.185\textwidth, height=0.185\textwidth]{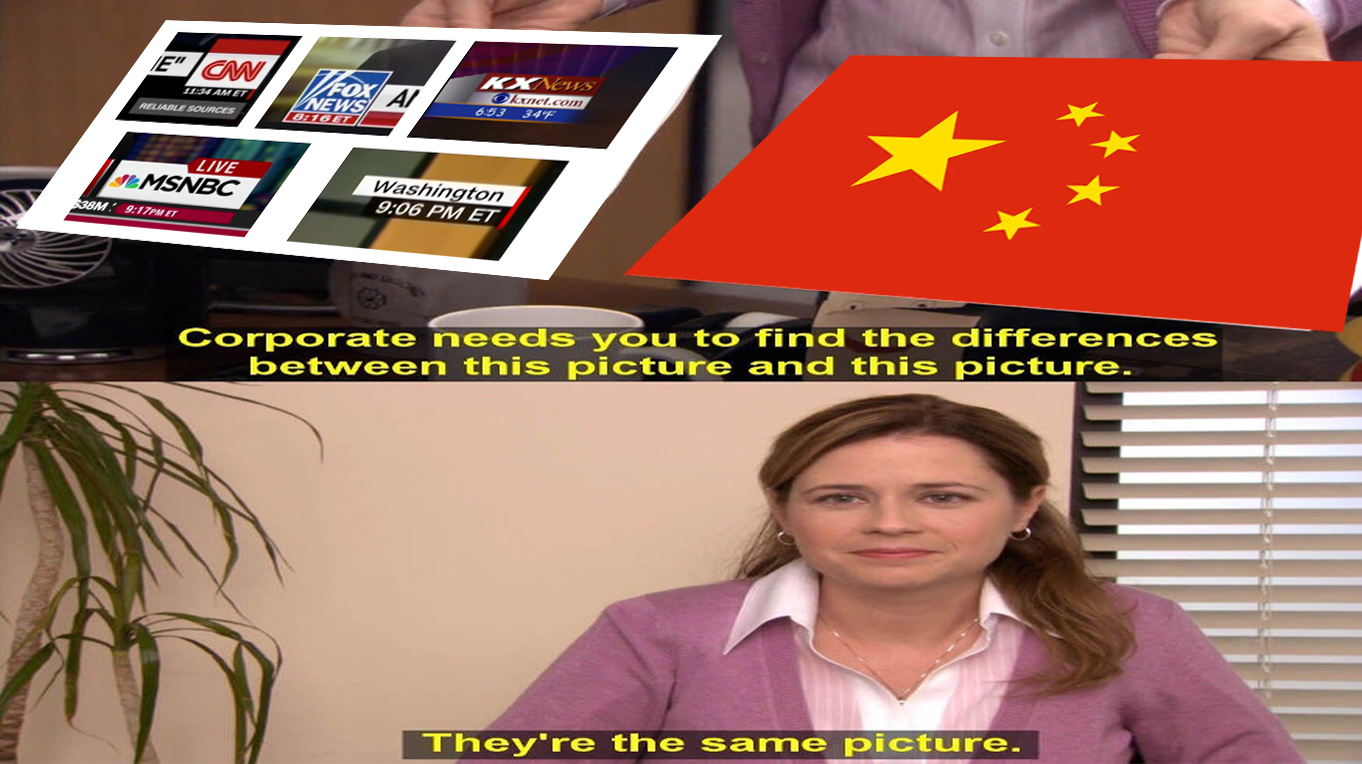}}\hspace{0.2em}
\subfloat[{\tt [2,2]}\label{fig:meme:exm:4}]{
\includegraphics[width=0.185\textwidth, height=0.185\textwidth]{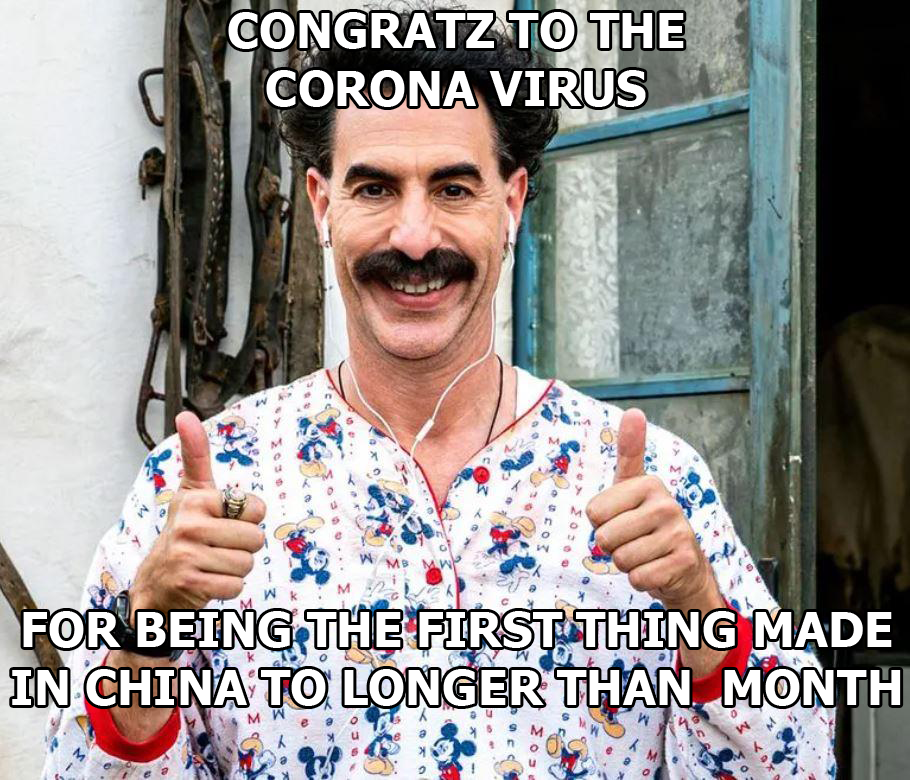}}\hspace{0.2em}
\subfloat[{\tt [2,3]}\label{fig:meme:exm:5}]{
\includegraphics[width=0.185\textwidth, height=0.185\textwidth]{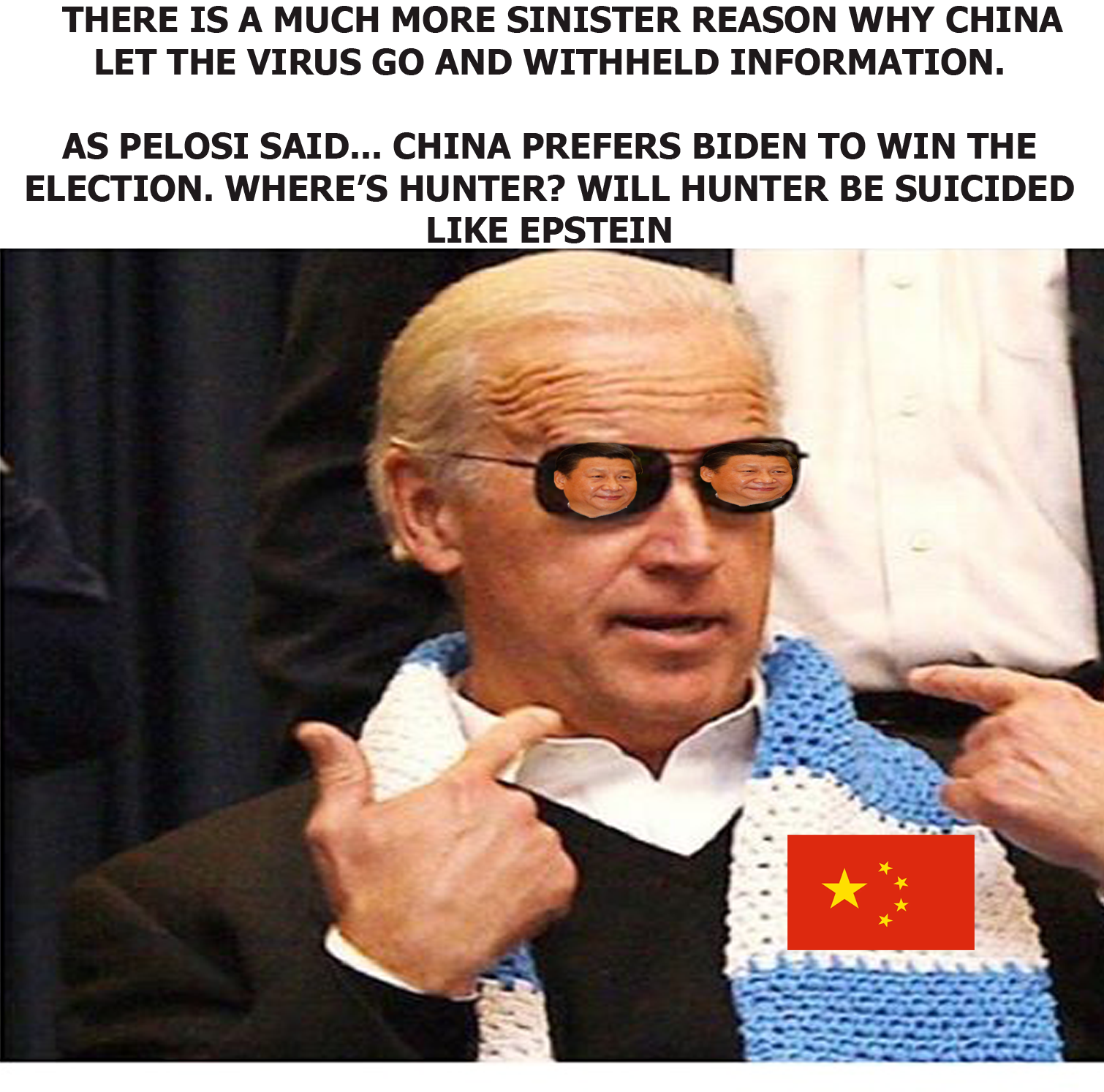}}\hspace{0.2em}

\caption{Examples from our \dataname\ dataset. The labels are in the format {\tt [Intensity, Target]}. For {\tt Intensity}, \{0, 1, 2\} correspond to {\em harmless}, {\em partially harmful}, and {\em very harmful}, respectively. For {\tt Target}, \{0, 1, 2, 3\} correspond to {\em individual}, {\em organization}, {\em community}, and {\em society}, respectively. Examples \ref{fig:meme:exm:2} and \ref{fig:meme:exm:3} are harmful, but neither hateful, nor offensive. Example~\ref{fig:meme:exm:4} is both harmful and offensive. \imgsrc{https://i.imgur.com/OeiN8Ou.png}{(a)}; \imgsrc{https://i.imgur.com/pJwKyfi.jpeg}{(b)}; \imgsrc{https://i.imgur.com/oZJM1W9.png}{(c) 1}, \imgsrc{https://i.imgur.com/ji4e16Z.jpeg}{(c) 2}, \imgsrc{https://i.imgur.com/zI29ADd.png}{(c) 3}; \imgsrc{https://i.imgur.com/i8PFAwg.jpeg}{(d)}; \imgsrc{https://i.imgur.com/RUicIXl.jpg}{(e) 1}, \imgsrc{https://i.imgur.com/zI29ADd.png}{(e) 2}, \imgsrc{https://search.creativecommons.org/photos/5756062d-7ddd-413c-be51-e18572a6ca38}{(e) 3}; \imgur{1}   \ccsnd{2}.
}
\label{fig:meme:example}
\end{figure*}

\section{Introduction}

The growing popularity of social media has led to the rise of multimodal content as a way to express ideas and emotions. As a result, a brand new type of message was born: \emph{meme}. A meme is typically formed by an image and a short piece of text on top of it, embedded as part of the image. Memes are typically innocent and designed to look funny.

Over time, memes started being used for harmful purposes in the context of contemporary political and socio-cultural events, targeting individuals, groups, businesses, and society as a whole. At the same time, their multimodal nature and often camouflaged semantics 
make their analysis highly challenging \cite{DBLP:journals/corr/abs-1910-02334}.

{\bf Meme analysis.} The proliferation of memes online and their increasing importance have led to a growing body of research on meme analysis \cite{sharma-etal-2020-semeval,reis2020dataset, pramanick2021exercise}. It has also been shown that off-the-shelf multimodal tools may be inadequate to unfold the underlying semantics of a meme as (\emph{i})~memes are often context-dependent, (\emph{ii})~the visual and the textual content are often uncorrelated,
and (\emph{iii}) meme images are mostly morphed, and the embedded text is sometimes hard to extract using standard OCR tools \cite{bonheme-grzes-2020-sesam}.

{\bf The dark side of memes.} Recently, there has been a lot of effort to explore the dark side of memes, e.g.,~focusing on hate \cite{kiela2020hateful} and offensive \cite{suryawanshi-etal-2020-multimodal} memes. However, the harm a meme can cause can be much broader. For instance, the meme\footnote{In order to avoid potential copyright issues, all memes we show in this paper are our own recreation of existing memes, using images with clear licenses.} in Figure~\ref{fig:meme:exm:3} is neither hateful nor offensive, but it is harmful to the media shown on the top left (ABC, CNN, etc.), as it compares them to China, suggesting that they adopt strong censorship policies. In short, the scope of harmful meme detection is {\em much broader}, and it may encompass other aspects such as cyberbullying, fake news, etc. Moreover, harmful memes have a target (e.g.,~news organization such as ABC and CNN in our previous example), which requires separate analysis not only to decipher their underlying semantics, but also to help with the explainability of the detection models. 

{\bf Our contributions.} In this paper, we study harmful memes, and we formulate two problems. {\bf Problem 1 (Harmful meme detection)}: Given a meme, detect whether it is {\em very harmful}, {\em partially harmful}, or {\em harmless}. {\bf Problem 2 (Target identification of harmful memes)}: Given a harmful meme, identify whether it targets an {\em individual}, an {\em organization}, a {\em community/country}, or the {\em society/general public/others}. To this end, we develop a novel dataset, \dataname, containing $3,544$ real memes related to COVID-19, which we collected from the web and carefully annotated. Figure~\ref{fig:meme:example} shows several examples of memes from our collection, whether they are harmful, as well as the types of their targets. We prepare detailed annotation guidelines for both tasks. We further experiment with ten state-of-the-art unimodal and multimodal models for benchmarking the two problems. Our experiments demonstrate that a systematic combination of multimodal signals is needed to tackle these problems.  Interpreting the models further reveals some of the biases that the best multimodal model exhibits, leading to the drop in performance. Finally, we argue that off-the-shelf models are inadequate in this context and that there is a need for specialized models

Our contributions can be summarized as follows: 
\begin{itemize}
 \item We study two new problems: (\emph{i})~detecting harmful memes and (\emph{ii})~detecting their targets. 
 \item We release a new benchmark dataset, \dataname, developed based on comprehensive annotation guidelines.
 \item We perform initial experiments with state-of-the-art textual, visual, and multimodal models to establish the baselines. We further discuss the limitations of these models.
 \end{itemize}

\noindent{\bf Reproducibility.} The full dataset and the source code of the baseline models are available at

\url{http://github.com/di-dimitrov/harmeme}

The appendix contains the values of the hyper-parameters and the detailed annotation guidelines.

\section{Related Work}
\label{sec:rel_work}

Below, we present an overview of the datasets and the methods used for multimodal meme analysis.

\textbf{Hate speech detection in memes.}
\citet{DBLP:journals/corr/abs-1910-02334} developed a collection of $5,020$ memes for hate speech detection.
Similarly, the Hateful Memes Challenge by Facebook introduced a dataset consisting of $10k+$ memes, annotated as hateful or non-hateful \cite{kiela2020hateful}. The memes were generated {\em artificially}, so that they resemble real ones shared on social media, along with ``benign confounders.'' As part of this challenge, an array of approaches with different architectures and features have been tried, including Visual BERT, ViLBERT, VLP, UNITER, LXMERT, VILLA, ERNIE-Vil, Oscar and other Transformers \cite{vaswani2017attention, li2019visualbert, zhou2019unified, tan2019lxmert, su2020vlbert, yu2020ernievil, li2020oscar, lippe2020multimodal, zhu2020enhance, muennighoff2020vilio}. Other approaches include multimodal feature augmentation and cross-modal attention mechanism using inferred image descriptions \cite{das2020detecting, sandulescu2020detecting, zhou2020multimodal, atri2021see}, as well as up-sampling confounders and loss re-weighting to complement multimodality \cite{lippe2020multimodal}, web entity detection along with fair face classification \cite{karkkainen2019fairface} from memes \cite{zhu2020enhance}, cross-validation ensemble learning and semi-supervised learning \cite{zhong2020classification} to improve robustness.

\textbf{Meme sentiment/emotion analysis.}
\citet{Hu_2018} developed the TUMBLR dataset for emotion analysis, consisting of image--text pairs along with associated tags, by collecting posts from the TUMBLR platform. \citet{duong2017multimodal} prepared a multimodal dataset containing images, titles, upvotes, downvotes, \#comments, etc., all collected from Reddit. Recently, SemEval-2020 Task 9 on Memotion Analysis \cite{sharma-etal-2020-semeval} introduced a dataset of $10k$ memes, annotated with sentiment, emotions, and emotion intensity.
Most participating systems in this challenge used fusion of visual and textual features computed using models such as Inception, ResNet, CNN, VGG-16 and DenseNet for image representation \cite{morishita-etal-2020-hitachi-semeval-2020, sharma-etal-2020-memebusters, yuan-etal-2020-ynu}, and BERT, XLNet, LSTM, GRU and DistilBERT for text representation \cite{liu-etal-2020-uor, gundapu-mamidi-2020-gundapusunil}. Due to class imbalance in the dataset, approaches such as GMM and Training Signal Annealing (TSA) were also found useful.  \citet{morishita-etal-2020-hitachi-semeval-2020, bonheme-grzes-2020-sesam, guo-etal-2020-guoym, sharma-etal-2020-memebusters} proposed ensemble learning, whereas \citet{gundapu-mamidi-2020-gundapusunil, de-la-pena-sarracen-etal-2020-prhlt} and several others used multimodal approaches. A few others leveraged transfer-learning using pre-trained models such as BERT \cite{devlin2019bert}, VGG-16 \cite{simonyan2014very}, and ResNet \cite{he2016deep}. Finally, state-of-the-art results for all three tasks ---sentiment classification, emotion classification and emotion quantification on this dataset,--- were reported by \citet{pramanick2021exercise}, who proposed a deep neural model that combines sentence demarcation and multi-hop attention. They also studied the interpretability of the model using the LIME framework \cite{ribeiro2016should}.

\textbf{Meme propagation.} \citet{9060100} surveyed personality traits of social media users who are more active in spreading misinformation in the form of memes. 
\citet{dis_meme} studied the characteristics of memes as a vehicle for spreading potential misinformation and disinformation. \citet{zannettou2019characterizing} discussed the quantitative aspects of large-scale dissemination of racist and hateful memes among polarized communities on platforms such as 4chan's /pol/.  \citet{ling2021dissecting} examined the artistic composition and the aesthetics of memes, the subjects they communicate, and the potential for virality. 
Based on this analysis, they manually annotated $50$ memes as viral vs. non-viral. \citet{zannettou2019quantitative} analyzed the ``Happy merchant'' memes and showed how online fringe communities influence their spread to mainstream social networking platforms. They reported reasonable agreement for most manually annotated labels, and established a characterization for meme virality.

\textbf{Other studies on memes.} 
\citet{reis2020dataset} built a dataset of memes related to the 2018 and the 2019 election in Brazil ($34k$ images, $17k$ users) and India ($810k$ images, $63k$ users) with focus on misinformation. 
Another dataset of 950 memes targeted the propaganda techniques used in memes \cite{ACL2021:propaganda:memes}, which was also featured as a shared that at SemEval-2021 \cite{SemEval2021:task6}.
\citet{10.1145/1557019.1557077} introduced a dataset of $96$ million memes collected from various links and blog posts between August 2008 and April 2009 for tracking the most frequently appearing stories, phrases, and information. 
Topic modeling of textual and visual cues of hate and racially abusive multi-modal content over sites such as 4chan was studied for scenarios that leverage genetic testing to claim superiority over minorities \cite{mittos2019and}.
\citet{zannettou2019characterizing} examined the content of meme images and online posting activities to identify the probability of occurrence of one event in a specific background process, affecting the occurrence of other events in the rest of the processes, also known as Hawkes process \cite{Hawkes1971SpectraOS}, within the context of online posting of trolls.  \citet{wang2020understanding} observed that fauxtographic content tends to attract more attention, and established how such content becomes a meme in social media. Finally, there is a recent survey on multi-modal disinformation detection \cite{Survey:2021:Multimodal:Disinformation}.

{\bf Differences with existing studies.} Hate speech detection in multimodal memes \cite{kiela2020hateful} is the closest work to ours. However, we are substantially different from it and from other related studies as (\emph{i})~we deal with \emph{harmful} meme detection, which is a more {\em general} problem than \emph{hateful} meme detection;  (\emph{ii})~along with harmful meme detection, we also identify the {\em entities that the harmful meme targets}; (\emph{iii})~our \dataname\ comprises {\em real-world memes} posted on the web as opposed to using synthetic memes as in \cite{kiela2020hateful}; and (\emph{iv})~we present a unique dataset and benchmark results for harmful meme detection and for identifying the target of harmful memes.

\section{\emph{Harmful} Meme: Definition}\label{sec:definition}
Here, we define {\em harmful memes} as follows: {\em multimodal units consisting of an image  and a piece of text embedded that has the potential to cause harm to an individual, an organization, a community, or the society more generally}. Here, \emph{harm} includes mental abuse, defamation, psycho-physiological injury, proprietary damage, emotional disturbance, and compensated public image.

{\bf Harmful vs. hateful/offensive.} \emph{Harmful} is a more general term than \emph{offensive} and \emph{hateful}: \emph{offensive} and \emph{hateful} memes are \emph{harmful}, but not all \emph{harmful} memes are \emph{offensive} or \emph{hateful}. For instance, the memes in Figures \ref{fig:meme:exm:2} and \ref{fig:meme:exm:3} are neither offensive nor hateful, but harmful to {\em Donald Trump} and to {\em news media} such as CNN, respectively.
Offensive memes typically aim to mock or to bully a social entity. A hateful meme contains offensive content that targets an entity (e.g.,~an individual, a community, or an organization) based on its personal/sensitive attributes such as gender, ethnicity, religion, nationality, sexual orientation, color, race, country of origin, and/or immigration status. The \emph{harmful} content in a harmful meme is often camouflaged and might require critical judgment to establish its potencial to do hard. Moreover, the social entities attacked or targeted by harmful memes can be any individual, organization, or community, as opposed to \emph{hateful} memes, where entities are attacked based on personal attributes.

\section{Dataset}
\label{sec:dataset}

Below, we describe the data collection, the annotation process and the guidelines, and we give detailed statistics about the \dataname\ dataset.

\subsection{Data Collection and Deduplication}
To collect potentially harmful memes in the context of COVID-19, we searched using different services, mainly \emph{Google Image Search}. We used keywords such as \textit{Wuhan Virus Memes}, \textit{US Election and COVID Memes}, \textit{COVID Vaccine Memes}, \textit{Work From Home Memes}, \textit{Trump Not Wearing Mask Memes}. We then used an extension\footnote{\url{http://download-all-images.mobilefirst.me/}} of Google Chrome to download the memes. We further scraped various publicly available groups on \textit{Instagram} for meme collection. Note that, adhering to the terms of social media, we did not use content from any private/restricted pages.

Unlike the Hateful Memes Challenge \citep{kiela2020hateful}, which used synthetically generated memes, our \dataname\ dataset contains {\em original memes} that were actually shared in social media. As all memes were gathered from real sources, we maintained strict filtering criteria\footnote{Details are given in Appendix \ref{sec:filtering}.} on the resolution of meme images and on the readability of the meme text during the collection process.
We ended up collecting $5,027$ memes. However, as we collected memes from independent sources, we had some duplicates. We thus used two efficient de-duplication repositories\footnote{\url{gitlab.com/opennota/findimagedupes}} \footnote{\url{http://github.com/arsenetar/dupeguru}} sequentially, and we preserved the memes with the highest resolution from each group of duplicates. We removed $1,483$ duplicate memes, thus ending up with a dataset of $3,544$. Although we tried to collect only harmful memes, the dataset contained memes with various levels of harmfulness, which we manually labeled during the annotation process, as discussed in Section~\ref{Annotation Process}. We further used Google's OCR Vision API\footnote{\url{http://cloud.google.com/vision}} to extract the textual content of each meme.

\begin{figure}[!t]
    \centering
    \includegraphics[width=\columnwidth]{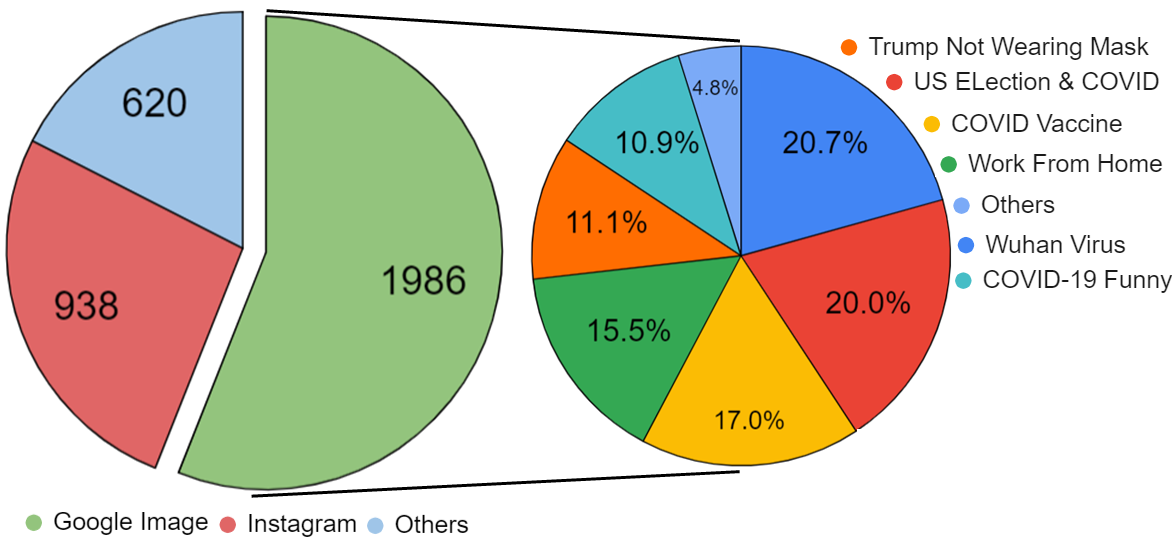}
    \caption{Statistics about the \dataname\ dataset. On the left, we show the distribution by source, while on the right, we show the percentage of memes collected by corresponding keywords in Google Image Search.}
    \label{fig:source_statistics}
\end{figure}

\subsection{Annotation Guidelines} \label{Annotation Guidelines}

As discussed in Section \ref{sec:definition}, we consider a meme as harmful only if it is implicitly or explicitly intended to cause \emph{harm} to an entity, depending on the personal, political, social, educational or industrial background of that entity. The intended \emph{harm} can be expressed in an obvious manner such as by abusing, offending, disrespecting, insulting, demeaning, or disregarding the entity or any sociocultural or political ideology, belief, principle, or doctrine associated with that entity. Likewise, the \emph{harm} can also be in the form of a more subtle attack such as mocking or ridiculing a person or an idea. 

We asked the annotators to label the intensity of the harm as \textit{harmful} or \textit{partially harmful}, depending upon the context and the ingrained explication of the meme. Moreover, we formally defined four different classes of targets and compiled well-defined guidelines\footnote{More details of the annotation guidelines are presented in Appendix \ref{sec:annotation}.} that the annotators adhered to while manually annotating the memes. The four target entities are as follows (c.f. Figure \ref{fig:meme:example}):   

\begin{enumerate}  
   \item \textbf{Individual:} A person, usually a celebrity (e.g.,~a well-known politician, an actor, an artist, a scientist, an environmentalist, etc. such as \emph{Donald Trump, Joe Biden, Vladimir Putin, Hillary Clinton, Barack Obama, Chuck Norris, Greta Thunberg, Michelle Obama}).  
    \item \textbf{Organization:} An organization is a group of people with a particular purpose, such as a business, a governmental department, a company, an institution or an association, comprising more than one person, and having a particular purpose, such as research organizations (e.g.,~\emph{WTO, Google}) and political organizations (e.g.,~\emph{the Democratic Party}).

    \item \textbf{Community:} A community is a social unit with commonalities based on personal, professional, social, cultural, or political attributes such as religious views, country of origin, gender identity, etc. Communities may share a sense of place situated in a given geographical area (e.g.,~a country, a village, a town, or a neighborhood) or in virtual space through communication platforms (e.g.,~online forums based on religion, country of origin, gender).
    \item \textbf{Society:} When a meme promotes conspiracies or hate crimes, it becomes harmful to the general public, i.e.,~to the entire society. 
\end{enumerate}
    
During the process of collection and annotation, we rejected memes based on the following four criteria:
    (\emph{i})~the meme text is in code-mixed or non-English language; (\emph{ii})~the meme text is not readable (e.g.,~blurry text, incomplete text, etc.); (\emph{iii})~the meme is unimodal, containing only textual or visual content; (\emph{iv})~the meme contains cartoons (we added this last criterion as cartoons can be hard to analyze by AI systems).

\begin{figure}[!t]
\centering
\begin{subfigure}[ht!]{.37\textwidth}
\centering
\includegraphics[width=1\textwidth]{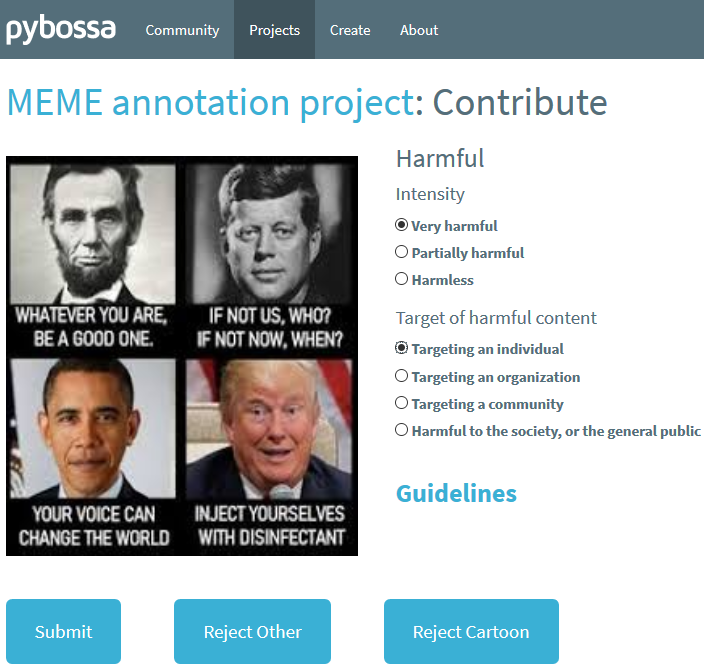}
\caption{Annotation interface}
\vspace{6pt}
\label{fig:annotation}
\end{subfigure}
\begin{subfigure}[ht!]{.37\textwidth}
\centering
\includegraphics[width=1\textwidth]{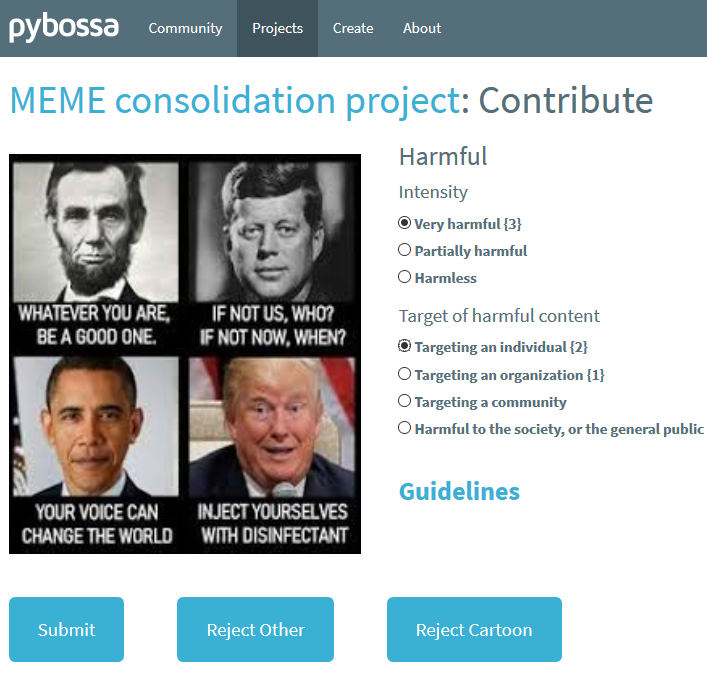}
\caption{Consolidation interface}
\label{fig:consolidation}
\end{subfigure}
\caption{Snapshot of the PyBossa GUI used for annotation and consolidation.}
\label{fig:platform}
\end{figure}

\subsection{Annotation Process} \label{Annotation Process}

For the annotation process, we had $15$ annotators, including professional linguists and researchers in Natural Language Processing (NLP): $10$ of them were male and the other $5$ were female, and their age ranged between $24$--$45$ years. We used the PyBossa\footnote{\url{http://pybossa.com/}} crowdsourcing framework for our annotations (c.f. Figure \ref{fig:platform}). We split the annotators into five groups of three people, and each group annotated a different subset of the data. Each annotator spent about $8.5$ minutes on average to annotate one meme. At first, we trained our annotators with the definition of harmful memes and their targets, along with the annotation guidelines. 
To achieve quality annotation, our main focus was to make sure that the annotators were able to understand well what harmful content is and how to differentiate it from humorous, satirical, hateful, and non-harmful content.

{\bf Dry run.}
We conducted a dry run on a subset of $200$ memes, which helped the annotators understand well the definitions of harmful memes and targets, as well as to eliminate the uncertainties about the annotation guidelines. Let $\alpha_i$ be a single annotator. For the preliminary data, we computed the inter-annotator agreement in terms of Cohen's $\kappa$ \cite{bobicev-sokolova-2017-inter} for three randomly chosen annotators $\alpha$\textsubscript{[1,2,3]} for each meme for both tasks. The results are shown in Table~\ref{tab:inter-anno}. We can see that the score is low for both tasks ($0.295$ and $0.373$), which is expected for the initial dry run. With the progression of the annotation phases, we observed much higher agreement, thus confirming that the dry run helped to train the annotators. 

{\bf Final annotation.} After the dry run, we started the final annotation process. Figure~\ref{fig:annotation} shows an example annotation of the PyBossa annotation platform. We asked the annotators to check whether a given meme falls under the four rejection criteria as given in the annotation guidelines. After confirming the validity of the meme, it was rated by three annotators for both tasks.

\textbf{Consolidation.} 
In the consolidation phase, for high agreements, we used majority voting to decide the final label, and we added a fourth annotator otherwise. Table~\ref{tab:dataset} shows statistics about the labels and the data splits.
After the final annotation, Cohen's $\kappa$ increased to $0.695$ and $0.797$ for the two tasks, which is moderate and high agreement, respectively. These scores show the difficulty and the variability in gauging the \textit{harmfulness} by human experts. For example, we found memes where two annotators independently chose \textit{partially harmful}, but the third annotator annotated it as \textit{very harmful}.

\begin{table}[!t]
\begin{center}

\resizebox{0.8\columnwidth}{!}{
\begin{tabular}{p{1.85cm}| p{1.8cm} | c c c } 
\hline
& \centering \bf Phase & \multicolumn{2}{c}{\bf Annotators} & $\kappa$ \\
\hline\hline
\multirow{6}{2cm}{\centering Harmful meme detection} & \multirow{3}{2cm}{\centering Trial Annotation} & $\alpha_1$ & $\alpha_2$ & 0.29  \\
\cline{3-5}
& & $\alpha_1$ & $\alpha_3$ & 0.34  \\
\cline{3-5}
& & $\alpha_2$ & $\alpha_3$ & 0.26 \\
\cline{2-5}
& \multirow{3}{*}{\parbox{2cm}{\centering Final Annotation}} & $\alpha_1$ & $\alpha_2$ &  0.67 \\ 
\cline{3-5}
& & $\alpha_1$ & $\alpha_3$ & 0.75  \\
\cline{3-5}
& & $\alpha_2$ & $\alpha_3$ & 0.72  \\

\hline
\hline 

\multirow{6}{2cm}{\centering Target identification} & \multirow{3}{2cm}{\centering Trial Annotation} & $\alpha_1$ & $\alpha_2$ & 0.35  \\
\cline{3-5}
& & $\alpha_1$ & $\alpha_3$ & 0.38 \\
\cline{3-5}
& & $\alpha_2$ & $\alpha_3$ & 0.39 \\
\cline{2-5}
& \multirow{3}{*}{\parbox{2cm}{\centering Final Annotation}} & $\alpha_1$ & $\alpha_2$ &  0.77 \\ 
\cline{3-5}
& & $\alpha_1$ & $\alpha_3$ & 0.83  \\
\cline{3-5}
& & $\alpha_2$ & $\alpha_3$ & 0.79 \\
 \hline
 \hline
\end{tabular}}
\end{center}
\caption{\label{tab:table-name} Cohen's $\kappa$ agreement during different phases of annotation for each task: harmful meme detection ($3$-class classification) and target identification ($4$-class classification) of harmful memes.}
\label{tab:inter-anno}
\end{table}

\subsection{Lexical Analysis of \dataname }\label{analysis}

\begin{figure*}[!t]
\centering
\includegraphics[width=1.6\columnwidth]{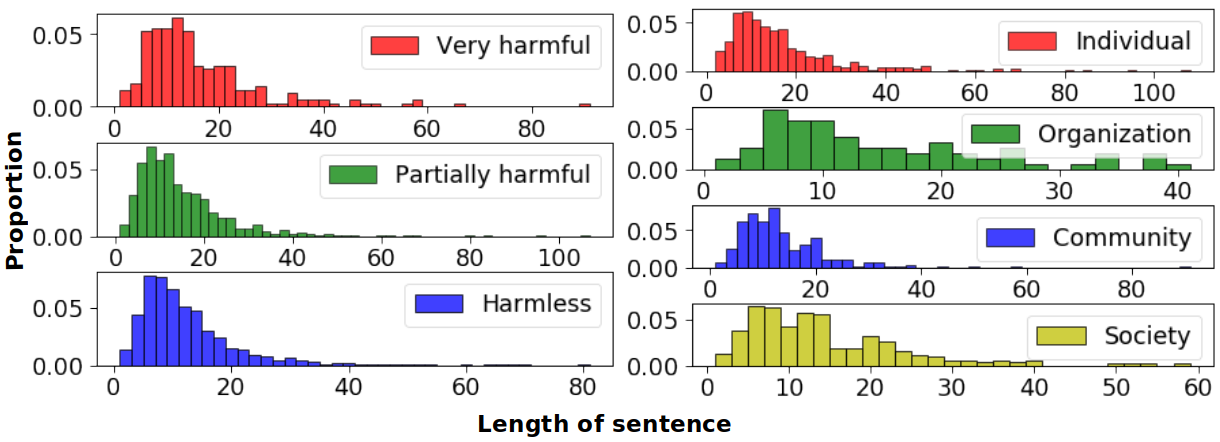}
\caption{Histogram of the length of the meme' text for each class: for harmfulness on the left, and for the target of harmful memes on the right.}
\label{fig:histogram}
\end{figure*}

Figure \ref{fig:histogram} shows the length distribution of the meme text for both tasks, and Table~\ref{tab:textual_lexical} shows the top-5 most frequent words in the union of the validation and the test sets. We can see that names of politicians and words related to COVID-19 are frequent in \textit{very harmful} and \textit{partially harmful} memes. For the target of the harmful memes, we notice the presence of various class-specific words such as \emph{president, trump, obama, china}. These words often incorporate bias in the machine learning models, which makes the dataset more challenging and difficult to learn from (see Section~\ref{sec:interpretation} for more detail).

\begin{table*}[t!]
\centering
    
\resizebox{\textwidth}{!}
{
\begin{tabular}{l|r|r r r|r|r r r r}
\hline
 & \multirow{2}{*}{\bf \#Memes} & \multicolumn{3}{c|}{\bf Harmfulness} & \multirow{2}{*}{\bf \#Memes} & \multicolumn{4}{c}{\bf Target}\\ \cline{3-5} \cline{7-10}
 
& & \bf Very Harmful & \bf Partially Harmful & \bf Harmless & & \bf Individual & \bf Organization & \bf Community & \bf Society\\ \hline 

\hline 
Train & 3,013 & 182 & 882 & 1,949 & 1,064 & 493 & 66 & 279 & 226\\
Validation & 177 & 10 & 51 & 116 & 61 & 29 & 3 & 16 & 13 \\ 
Test & 354 & 21 & 103 & 230 & 124 & 59 & 7 & 32 & 26 \\
\hline
\textbf{Total} & 3,544 & 213 & 1,036 & 2,295 & 1,249 & 582 & 75 & 327 & 265 \\

\hline
\end{tabular}}

\caption{Statistics about the \dataname\  dataset. The memes belonging to the \textit{very harmful} and the \textit{partially harmful} categories are annotated with one of the following four targets: \textit{individual}, \textit{organization}, \textit{community}, or \textit{society}. }

\label{tab:dataset}
%\vspace{-2mm}
\end{table*}

\begin{table*}[t!]
\centering
    
\resizebox{0.99\textwidth}{!}
{
\begin{tabular}{c | c | c || c | c | c | c}
\hline
\multicolumn{3}{c||}{\bf Harmfulness} & \multicolumn{4}{c}{\bf Target}\\ 
\hline
\bf Very Harmful & \bf Partially Harmful & \bf Harmless & \bf Individual & \bf Organization & \bf Community & \bf Society\\ 

\hline

mask (\textcolor{black}{0.0512}) & trump (\textcolor{black}{0.0642}) & you (\textcolor{black}{0.0264}) & trump (\textcolor{black}{0.0541}) & deadline (\textcolor{black}{0.0709}) & china (\textcolor{black}{0.0665}) & mask (\textcolor{black}{0.0441}) \\

trump (\textcolor{black}{0.0404}) &  president (\textcolor{black}{0.0273}) & home (\textcolor{black}{0.0263}) & president (\textcolor{black}{0.0263}) & associated (\textcolor{black}{0.0709}) & chinese (\textcolor{black}{0.0417}) & vaccine	(\textcolor{black}{0.0430})\\

wear (\textcolor{black}{0.0385}) & obama (\textcolor{black}{0.0262}) & corona (\textcolor{black}{0.0251}) & donald (\textcolor{black}{0.0231}) & extra (\textcolor{black}{0.0645}) & virus (\textcolor{black}{0.0361}) & alcohol (\textcolor{black}{0.0309})
\\

thinks (\textcolor{black}{0.0308} & donald (\textcolor{black}{0.0241}) & work (\textcolor{black}{0.0222}) & obama (\textcolor{black}{0.0217}) & ensure (\textcolor{black}{0.0645}) & wuhan (\textcolor{black}{0.0359}) & temperatures (\textcolor{black}{0.0309}) \\

killed (\textcolor{black}{0.0269}) & virus (\textcolor{black}{0.0213}) & day (\textcolor{black}{0.0188}) &
covid (\textcolor{black}{0.0203}) & qanon (\textcolor{black}{0.0600}) & cases (\textcolor{black}{0.0319}) & killed (\textcolor{black}{0.0271})\\

\if 0
people (\textcolor{black}{0.0265}) & covid (\textcolor{black}{0.0207}) & when (\textcolor{black}{0.0185}) & usa (\textcolor{black}{0.0200}) & twitter (\textcolor{black}{0.0474}) & safe (\textcolor{black}{0.0279}) & people (\textcolor{black}{0.0264})\\

eliminate (\textcolor{black}{0.0245}) & memes (\textcolor{black}{0.0185}) & virus (\textcolor{black}{0.0177}) & biden (\textcolor{black}{0.0174}) & mask (\textcolor{black}{0.0384}) & coronavirus (\textcolor{black}{0.0221}) & coronavirus (\textcolor{black}{0.0261})\\

patriotic (\textcolor{black}{0.0245}) & biden (\textcolor{black}{0.0184}) & vaccine (\textcolor{black}{0.0165}) & positive (\textcolor{black}{0.0155}) & postive (\textcolor{black}{0.0354}) &
idiot (\textcolor{black}{0.0194}) & release (\textcolor{black}{0.0247})\\

medical	(\textcolor{black}{0.0228}) & usa (\textcolor{black}{0.0184}) & quarantine (\textcolor{black}{0.0161}) & twitter (\textcolor{black}{0.0143}) & guard (\textcolor{black}{0.0354}) &
government (\textcolor{black}{0.0174}) & just (\textcolor{black}{0.0235})\\

saddam (\textcolor{black}{0.0228}) & covfefe (\textcolor{black}{0.0158}) & working (\textcolor{black}{0.0154}) & emergency (\textcolor{black}{0.0135}) & chronology (\textcolor{black}{0.0354}) & picture (\textcolor{black}{0.0172}) & virus (\textcolor{black}{0.0225})\\
\fi
\hline
\end{tabular}}

\caption{Top-5 most frequent words per class. The tf-idf score per word  is given within parenthesis.}

\label{tab:textual_lexical}
%\vspace{-2mm}
\end{table*}

\section{Benchmarking \dataname\ dataset}
\label{sec:benchmark}
We provide benchmark evaluations on \dataname\ with a variety of state-of-the-art unimodal textual models, unimodal visual models, and models using both modalities. Except for unimodal visual models, we use MMF (Multimodal Framework)\footnote{\url{github.com/facebookresearch/mmf}} to conduct the necessary experiments. 

\subsection{Unimodal Models}
$\rhd$ \textbf{Text BERT:} We use textual BERT \cite{devlin2019bert} as the unimodal text-only model. \\
$\rhd$ \textbf{VGG19, DenseNet, ResNet, ResNeXt:} For the unimodal visual-only models, we used four different well-known models -- VGG19 \cite{simonyan2014very}, DenseNet-161 \citep{huang2017densely}, ResNet-152 \citep{he2016deep}, and ResNeXt-101 \citep{xie2017aggregated} pre-trained on the ImageNet \citep{deng2009imagenet} dataset. We extracted the feature maps from the last pooling layer of each architecture and fed them to a fully connected layer.

\subsection{Multimodal Models}

\noindent $\rhd$ \textbf{Late Fusion:} This model uses the mean score of pre-trained unimodal ResNet-152 and BERT. 
    
\noindent $\rhd$ \textbf{Concat BERT:} It concatenates the features extracted by pre-trained unimodal ResNet-152 and text BERT, and uses a simple MLP as the classifier. 
    
\noindent $\rhd$ \textbf{MMBT:} Supervised Multimodal Bitransformers   \citep{kiela2019supervised} is a multimodal architecture that inherently captures the intra-modal and the inter-modal dynamics within various input modalities. 
    
\noindent $\rhd$ \textbf{ViLBERT CC:} Vision and Language BERT (ViLBERT) \citep{lu2019vilbert}, trained on an intermediate multimodal objective (Conceptual Captions) \citep{sharma2018conceptual}, is a strong model with task-agnostic joint representation of image + text.
    
\noindent $\rhd$ \textbf{Visual BERT COCO:} Visual BERT (V-BERT) \citep{li2019visualbert} pre-trained on the multimodal COCO dataset \citep{lin2014microsoft} is another strong multimodal model used for a broad range of vision and language tasks.

%%%%%%%
% Two Single Column Tables
%%%%%%%
% First
%%%%%%%

\begin{table*}[hbtp]
\centering % centering table

\resizebox{\textwidth}{!}
{
\begin{tabular}{l | l | c c c c c c | c c c c c c}
\hline
\multirow{3}{*}{\bf Modality} & \multirow{3}{*}{\bf Model} & \multicolumn{12}{c}{\bf Harmful Meme Detection} \\ \cline{3-14}

& &  \multicolumn{6}{c|}{\bf $2$-Class Classification} & \multicolumn{6}{c}{\bf $3$-Class Classification} \\ \cline{3-14}

& & \bf Acc $\uparrow$ & \bf P $\uparrow$ & \bf R $\uparrow$& \bf F1 $\uparrow$ & \bf MAE $\downarrow$& \bf MMAE $\downarrow$& \bf Acc $\uparrow$& \bf P $\uparrow$& \bf R $\uparrow$& \bf F1 $\uparrow$& \bf MAE $\downarrow$& \bf MMAE $\downarrow$ \\

\hline
\hline

& Human$^\dagger$ & 90.68 & 84.35 & 84.19 & 83.55 & 0.1760 & 0.1723 & 86.10 & 67.35 & 65.84 & 65.10 & 0.2484 & 0.4857 \\ 

 & Majority & 64.76 & 32.38 & 50.00 & 39.30 & 0.3524 & 0.5000 &64.76 & 21.58 & 33.33 & 26.20 & 0.4125& 1.0 \\
\hline

\multirow{1}{4cm}{{\tt Text} Only} & TextBERT & 70.17 & 65.96 & 66.38& 66.25 &  0.3173 &0.2911 & 68.93 & 48.49 & 49.15 & 48.72 & 0.3250 & 0.5591 \\

\hline

\multirow{4}{4cm}{{\tt Image} Only} & VGG19 & 68.12 & 60.25 & 61.23 & 61.86 & 0.3204 & 0.3190 & 66.24 & 40.95 & 44.02 & 41.76 & 0.3198 & 0.6487 \\
& DenseNet-161 & 68.42 & 61.08 & 62.10 & 62.54 & 0.3202 & 0.3125 & 65.21 & 41.88 & 44.25 & 42.15 & 0.3102 & 0.6326\\
& ResNet-152 & 68.74 & 61.86 & 62.89 & 62.97 & 0.3188 & 0.3114 & 65.29 & 41.95 & 44.32 & 43.02 & 0.3047 & 0.6264\\
& ResNeXt-101 & 69.79 & 62.32 & 63.26 & 63.68 & 0.3175 & 0.3029 & 66.55 & 42.62 & 44.87 & 43.68 & 0.3036 & 0.6499 \\

\hline

\multirow{3}{4cm}{{\tt Image} + {\tt Text} (Unimodal Pre-training)} & Late Fusion & 73.24 & 70.28 & 70.36 &70.25 & 0.3167 & 0.2927 & 66.67 & 44.96 & 50.02 & 45.06 & 0.3850 & 0.6077 \\
& Concat BERT & 71.82& 71.58 & 72.23 & 71.82 & 0.3033  & 0.3156  & 65.54 & 42.29 & 45.42 & 43.37 & 0.3881 & 0.5976 \\
& MMBT & 73.48 & 68.89 & 68.95 & 67.12 & 0.3101 & 0.3258 & 68.08 & 51.72 & 51.94 & 50.88 & 0.3403 & 0.6474 \\

\hline

\multirow{2}{4cm}{{\tt Image} + {\tt Text} (Multimodal Pre-training)}
& ViLBERT CC & 78.53 & 78.62 &\textbf{81.41} &78.06 &0.2279 & 0.1881 & \textbf{75.71} & 48.89 & 49.21 & 48.82 & \textbf{0.2763} & 0.5329 \\
& V-BERT COCO & \textbf{81.36} & \textbf{79.55} & 81.19 & \textbf{80.13} & \textbf{0.1972} & \textbf{0.1857} & 74.01 &\textbf{ 56.35} & \textbf{54.79} & \textbf{53.85} & 0.3063 & \textbf{0.5303} \\

\hline
\end{tabular}}
\caption{Performance for harmful meme detection. For two-class classification, we merge \textit{very harmful} and \textit{partially harmful} into a single class. $^\dagger$ This row reports the human accuracy on the test set.}
\label{tab:results_hateful}
\end{table*}

%%%%%%%%
% Second
%%%%%%%%

\begin{table}[hbtp]
\centering % centering table

\resizebox{0.99\columnwidth}{!}
{
\begin{tabular}{l | l | c c c c c c}
\hline
\multirow{2}{*}{\bf Modality} & \multirow{2}{*}{\bf Model} & \multicolumn{6}{c}{\bf Target Identification of Harmful Memes} \\ \cline{3-8}

& & \bf Acc $\uparrow$ & \bf P $\uparrow$ & \bf R $\uparrow$& \bf F1 $\uparrow$ & \bf MAE $\downarrow$& \bf MMAE $\downarrow$ \\

\hline
\hline
& Human$^\dagger$ & 87.55 & 82.28 & 84.15 & 82.01 & 0.7866 & 0.3647 \\
  & Majority & 46.60 & 11.65 & 25.00 & 15.89 & 1.2201 & 1.5000 \\

\hline

\multirow{1}{3cm}{Text (T) only} & TextBERT & 69.35 & 55.60 & 54.37 & 55.60 & 1.1612 & 0.8988\\

\hline

\multirow{4}{3cm}{Image (I) only} & VGG19 & 63.48 & 53.85 & 54.02 & 53.60 & 1.1687 & 1.0549 \\
& DenseNet-161 & 64.52 & 53.96 & 53.95 & 53.51 & 1.1655 & 1.0065\\
& ResNet-152 & 65.75 & 54.25 & 54.13 & 53.78 & 1.1628 & 1.0459\\
& ResNeXt-101 & 65.82 & 54.47 & 54.20 & 53.95 & 1.1616 & 0.9277\\

\hline

\multirow{3}{3cm}{I + T (Unimmodal Pre-training)} & Late Fusion & 72.58 & 58.43 & 58.83 & 58.43 & 1.1476 & 0.6318  \\ 
& Concat BERT & 67.74 & 54.79 & 49.65 & 49.77 & 1.1377 & 0.8879\\
& MMBT & 72.58 & 58.43 & 58.83 & 58.35 & 1.1476 & 0.6318\\

\hline

\multirow{2}{3cm}{I + T (Multimodal Pre-training)}
& ViLBERT CC & 72.58 & 59.92 & 55.78 & 57.17 & 1.1671 & 0.8035\\
& V-BERT COCO & \textbf{75.81} & \textbf{66.29} & \textbf{69.09} & \textbf{65.77} & \textbf{1.1078} & \textbf{0.5036}\\

\hline
\end{tabular}}
\caption{Performance for target identification of harmful memes ($^\dagger$human accuracy on the test set).}
\label{tab:results_target}
\end{table}

\section{Experimental Results}

Below, we report the performance of the models described in the previous section for each of the two tasks. We further discuss some biases that negatively impact performance. Appendix~\ref{sec:hyperparameters} gives additional details about training and the values of the hyper-parameters we used in our experiments. 

\paragraph{Evaluation measures}
We used six evaluation measures: Accuracy, Precision, Recall, Macro-averaged F1, Mean Absolute Error (MAE), and Macro-Averaged Mean Absolute Error (MMAE) \citep{baccianella2009evaluation}. For the first four measures, higher values are better, while for the last two, lower values are better. Since the test set is imbalanced, measures like macro F1 and MMAE are more relevant.

\subsection{Harmful Meme Detection}

Table \ref{tab:results_hateful} shows the results for the harmful meme detection task. We start our experiments by merging the \textit{very hateful} and the \textit{partially hateful} classes, thus turning the problem into an easier {\em binary classification}. Afterwards, we perform the 3-class classification task. Since the test set is imbalanced, the majority class baseline achieves $64.76\%$ accuracy. We observe that the unimodal visual models perform only marginally better than the majority class baseline, which indicates that they are insufficient to learn the underlying semantics of the memes.

Moving down the table, we see that the unimodal text model is marginally better than the visual models. Then, for multimodal models, the performance improves noticeably, and more sophisticated fusion techniques yield better results. We also notice the effectiveness of multimodal pre-training over unimodal pre-training, which supports the recent findings by \citet{singh2020we}. While both ViLBERT CC and V-BERT COCO perform similarly, the latter achieves better Macro F1 and MMAE, which are the most relevant measures.

\subsection{Target Identification for Harmful Memes}

Table \ref{tab:results_target} shows the results for the target identification task. This is an imbalanced 4-class classification problem, and the majority class baseline yields $46.60\%$ accuracy. The unimodal models perform relatively better here, achieving $63\%-70\%$ accuracy; their F1 Macro and MMAE scores are also above the majority class. However, the overall performance of the unimodal models is poor. Incorporating multimodal signals with fine-grained fusion improves the results substantially, and advanced multimodal fusion techniques with multimodal pre-training perform much better than simple late fusion with unimodal pre-training. Moreover, V-BERT COCO outperforms ViLBERT CC by $8\%$ of F1 score and by nearly $0.3$ of MMAE.

\subsection{Human Evaluation}
To understand how human subjects perceive these tasks, we further hired a different set of experts (not the annotators) to label the test set. We observed $86\%-91\%$ accuracy on average for both tasks, which is much higher than V-BERT, the best-performing model. This shows that their is a potential for enriched multimodal models that better understand the ingrained semantics of the memes.

\begin{figure}[t]
\centering
\subfloat[{Very harmful meme}\label{fig:input_meme_harmful}]{
\includegraphics[width=0.207\textwidth, height=0.20\textwidth]{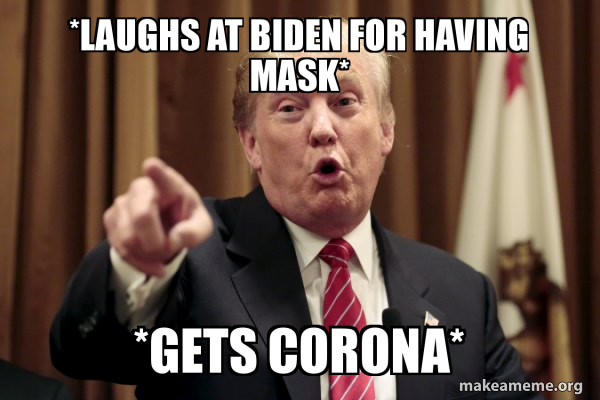}}
\hspace{2em}
\subfloat[{LIME output - image}\vspace{9pt}\label{fig:lime_image_harmful}]{
\includegraphics[width=0.207\textwidth, height=0.205\textwidth]{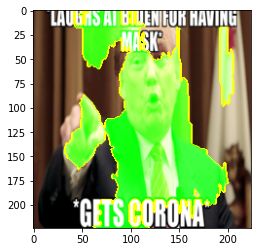}}
\\
\subfloat[{LIME output - text}
\vspace{9pt}\label{fig:lime_txt_harmful}]{
\includegraphics[width=\columnwidth]{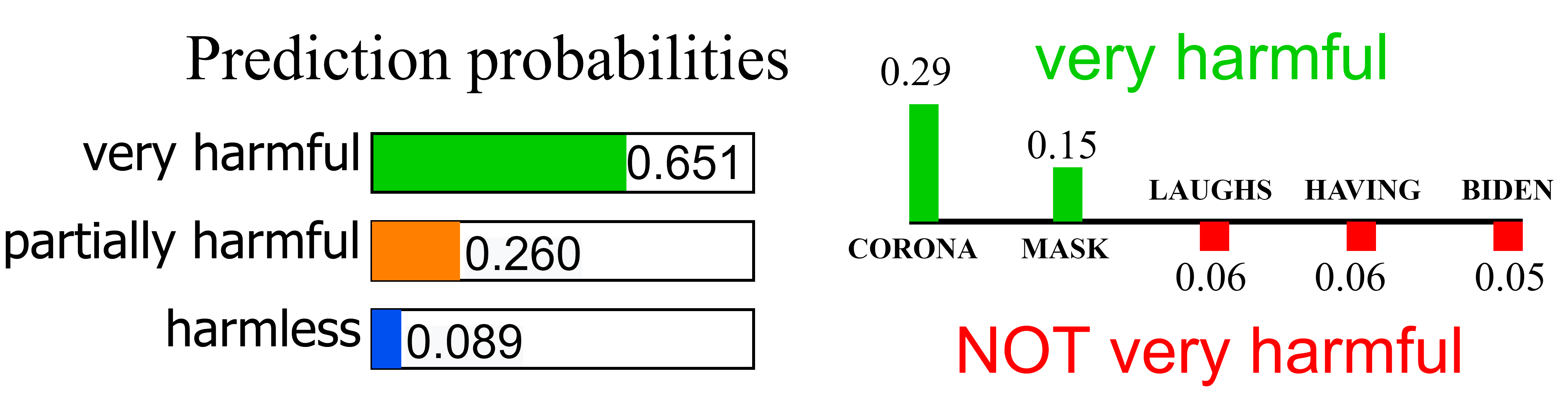}}
\\
\subfloat[{Harmless meme} \label{fig:input_meme_not_harmful}]{
\includegraphics[width=0.207\textwidth,height=0.205\textwidth]{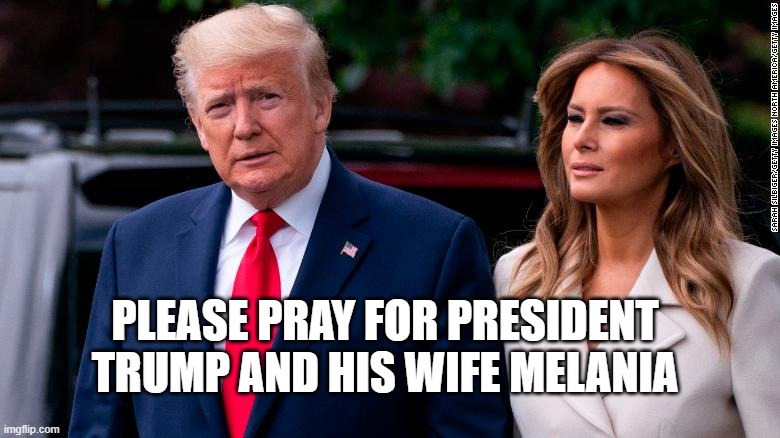}}\hspace{2em}
\subfloat[{LIME output - image}\label{fig:lime_image_not_harmful}]{
\includegraphics[width=0.207\textwidth, height=0.205\textwidth]{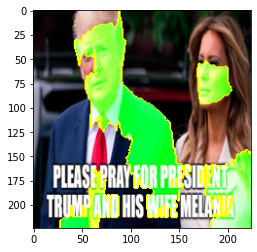}}
\caption{Example of explanation by LIME on both visual and textual modalities and visualization of bias in V-BERT for both tasks.}
\label{fig:lime}
\end{figure}

\subsection{Side-by-side Diagnostics and Anecdotes}\label{sec:interpretation}

Since the \dataname\ dataset was compiled of memes related to COVID-19, we expected that models with enriched contextual knowledge and sophisticated technique would have superior performance. Thus, to comprehend the interpretability of V-BERT (the best model), we used LIME (Locally Interpretable Model-Agnostic Explanations) 
\citep{ribeiro2016should}, a consistent model-agnostic explainer to interpret the predictions.

We chose two memes from the test set to analyze the potential explanability of V-BERT. The first meme, which is shown in Figure~\ref{fig:input_meme_harmful}, was manually labeled as \textit{very harmful}, and V-BERT successfully classified it, with prediction probabilities of $0.651$, $0.260$, and $0.089$ corresponding to the {\em very harmful}, the {\em partially harmful}, and the {\em harmless} classes respectively. Figure~\ref{fig:lime_image_harmful} highlights the most contributing super-pixels to the {\em very harmful} (green) class. As expected, the face of Donald Trump, as highlighted by the green pixels, prominently contributed to the prediction. Figure~\ref{fig:lime_txt_harmful} demonstrates the contribution of different meme words to the model prediction. We can see that words like \emph{CORONA} and \emph{MASK} have significant contributions to the {\em very harmful} class, thus supporting the lexical analysis of \dataname\ as shown in Table~\ref{tab:textual_lexical}. 

The second meme, which is shown in Figure~\ref{fig:input_meme_not_harmful}, was manually labeled as \textit{harmless}, but V-BERT  incorrectly predicted it to be {\em very harmful}. Figure~\ref{fig:lime_image_not_harmful} shows that, similarly to the previous example, the face of Donald Trump contributed to the prediction of the model. We looked closer into our dataset, and we found that it contained many memes with the image of Donald Trump, and that the majority of these memes fall under the {\em very harmful} category and targeted and individual. Therefore, instead of leaning the underlying semantics of one particular meme, the model easily got biased by the presence of Donald Trump's image and blindly classified the meme as {\em very harmful}. 

\section{Conclusion and Future Work}
\label{sec:conc_future}

We presented \dataname, the first large-scale benchmark dataset, containing 3,544 memes, related to COVID-19, with annotations for degree of harmfulness ({\em very harmful}, {\em partially harmful}, or {\em harmless}), 
as well as for the target of the harm (an {\em individual}, an {\em organization}, a {\em community}, or {\em society}). The evaluation results using several unimodal and multimodal models highlighted the importance of modeling the multimodal signal (for both tasks) ---(\emph{i})~detecting harmful memes and (\emph{ii})~detecting their targets---, 
and indicated the need for more sophisticated methods. 
We also analyzed the best model and identified its limitations.

In future work, we plan to design new multimodal models and to extend \dataname\ with examples from other topics, as well as to other languages.
Alleviating the biases in the dataset and in the models are other important research directions.

\section*{Ethics and Broader Impact}

\paragraph{User Privacy.}

Our dataset only includes memes and it does not contain any user information.

\paragraph{Biases.}

Any biases found in the dataset are unintentional, and we do not intend to do harm to any group or individual. We note that determining whether a meme is harmful can be subjective, and thus it is inevitable that there would be biases in our gold-labeled data or in the label distribution. We address these concerns by collecting examples using general keywords about COVID-19, and also by following a well-defined schema, which sets explicit definitions during annotation. Our high inter-annotator agreement makes us confident that the assignment of the schema to the data is correct most of the time.

\paragraph{Misuse Potential.}

We ask researchers to be aware that our dataset can be maliciously used to unfairly moderate memes based on biases that may or may not be related to demographics and other information within the text. Intervention with human moderation would be required in order to ensure that this does not occur.

\paragraph{Intended Use.}

We present our dataset to encourage research in studying harmful memes on the web. We distribute the dataset for research purposes only, without a license for commercial use. We believe that it represents a useful resource when used in the appropriate manner.

\paragraph{Environmental Impact.}

Finally, we would also like to warn that the use of large-scale Transformers requires a lot of computations and the use of GPUs/TPUs for training, which contributes to global warming \cite{strubell-etal-2019-energy}. This is a bit less of an issue in our case, as we do not train such models from scratch; rather, we fine-tune them on relatively small datasets. Moreover, running on a CPU for inference, once the model has been fine-tuned, is perfectly feasible, and CPUs contribute much less to global warming.

\section*{Acknowledgments}
The work was partially supported by the Wipro research grant and the Infosys Centre for AI, IIIT Delhi, India. 
It is also part of the Tanbih mega-project, developed at the Qatar Computing Research Institute, HBKU, which aims to limit the impact of ``fake news,'' propaganda, and media bias by making users aware of what they are reading.

\bibliographystyle{acl_natbib}
\bibliography{anthology_V2_xtinfo,acl2021}

\begin{thebibliography}{60}
\expandafter\ifx\csname natexlab\endcsname\relax\def\natexlab#1{#1}\fi

\bibitem[{Alam et~al.(2021)Alam, Cresci, Chakraborty, Silvestri, Dimitrov,
  Martino, Shaar, Firooz, and Nakov}]{Survey:2021:Multimodal:Disinformation}
Firoj Alam, Stefano Cresci, Tanmoy Chakraborty, Fabrizio Silvestri, Dimiter
  Dimitrov, Giovanni Da~San Martino, Shaden Shaar, Hamed Firooz, and Preslav
  Nakov. 2021.
\newblock A survey on multimodal disinformation detection.
\newblock \emph{arXiv 2103.12541}.

\bibitem[{Atri et~al.(2021)Atri, Pramanick, Goyal, and
  Chakraborty}]{atri2021see}
Yash~Kumar Atri, Shraman Pramanick, Vikram Goyal, and Tanmoy Chakraborty. 2021.
\newblock See, hear, read: Leveraging multimodality with guided attention for
  abstractive text summarization.
\newblock \emph{Knowledge-Based Systems}, page 107152.

\bibitem[{Baccianella et~al.(2009)Baccianella, Esuli, and
  Sebastiani}]{baccianella2009evaluation}
Stefano Baccianella, Andrea Esuli, and Fabrizio Sebastiani. 2009.
\newblock {E}valuation {M}easures for {O}rdinal {R}egression.
\newblock In \emph{Proceedings of the Ninth International Conference on
  Intelligent Systems Design and Applications}, ISDA~'09, pages 283--287, Pisa,
  Italy.

\bibitem[{Bobicev and Sokolova(2017)}]{bobicev-sokolova-2017-inter}
Victoria Bobicev and Marina Sokolova. 2017.
\newblock Inter-annotator agreement in sentiment analysis: Machine learning
  perspective.
\newblock In \emph{Proceedings of the International Conference Recent Advances
  in Natural Language Processing}, RANLP~'17, pages 97--102, Varna, Bulgaria.

\bibitem[{Bonheme and Grzes(2020)}]{bonheme-grzes-2020-sesam}
Lisa Bonheme and Marek Grzes. 2020.
\newblock {SESAM} at {S}em{E}val-2020 task 8: Investigating the relationship
  between image and text in sentiment analysis of memes.
\newblock In \emph{Proceedings of the Fourteenth Workshop on Semantic
  Evaluation}, SemEval~'20, pages 804--816, Barcelona, Spain.

\bibitem[{Crovitz and Moran(2020)}]{dis_meme}
Darren Crovitz and Clarice Moran. 2020.
\newblock {A}nalyzing {D}isruptive {M}emes in an {A}ge of {I}nternational
  {I}nterference.
\newblock \emph{The English Journal}, 109(4):62--69.

\bibitem[{Das et~al.(2020)Das, Wahi, and Li}]{das2020detecting}
Abhishek Das, Japsimar~Singh Wahi, and Siyao Li. 2020.
\newblock Detecting hate speech in multi-modal memes.
\newblock \emph{arXiv 2012.14891}.

\bibitem[{Deng et~al.(2009)Deng, Dong, Socher, Li, Li, and
  Fei-Fei}]{deng2009imagenet}
Jia Deng, Wei Dong, Richard Socher, Li-Jia Li, Kai Li, and Li~Fei-Fei. 2009.
\newblock {ImageNet}: A large-scale hierarchical image database.
\newblock In \emph{Proceedings of the IEEE Conference on Computer Vision and
  Pattern Recognition}, CVPR~'09, pages 248--255, Miami, FL, USA.

\bibitem[{Devlin et~al.(2019)Devlin, Chang, Lee, and
  Toutanova}]{devlin2019bert}
Jacob Devlin, Ming-Wei Chang, Kenton Lee, and Kristina Toutanova. 2019.
\newblock {BERT}: Pre-training of deep bidirectional transformers for language
  understanding.
\newblock In \emph{Proceedings of the 2019 Conference of the North {A}merican
  Chapter of the Association for Computational Linguistics: Human Language
  Technologies}, NAACL-HLT~'19, pages 4171--4186, Minneapolis, MN, USA.

\bibitem[{Dimitrov et~al.(2021{\natexlab{a}})Dimitrov, Bin~Ali, Shaar, Alam,
  Silvestri, Firooz, Nakov, and Da~San~Martino}]{ACL2021:propaganda:memes}
Dimitar Dimitrov, Bishr Bin~Ali, Shaden Shaar, Firoj Alam, Fabrizio Silvestri,
  Hamed Firooz, Preslav Nakov, and Giovanni Da~San~Martino. 2021{\natexlab{a}}.
\newblock Detecting propaganda techniques in memes.
\newblock In \emph{Proceedings of the Joint Conference of the 59th Annual
  Meeting of the Association for Computational Linguistics and the 11th
  International Joint Conference on Natural Language Processing},
  ACL-IJCNLP~'21.

\bibitem[{Dimitrov et~al.(2021{\natexlab{b}})Dimitrov, Bin~Ali, Shaar, Alam,
  Silvestri, Firooz, Nakov, and Da~San~Martino}]{SemEval2021:task6}
Dimitar Dimitrov, Bishr Bin~Ali, Shaden Shaar, Firoj Alam, Fabrizio Silvestri,
  Hamed Firooz, Preslav Nakov, and Giovanni Da~San~Martino. 2021{\natexlab{b}}.
\newblock {SemEval-2021 Task 6}: Detection of persuasion techniques in texts
  and images.
\newblock In \emph{Proceedings of the International Workshop on Semantic
  Evaluation}, SemEval~'21.

\bibitem[{Dupuis and Williams(2019)}]{9060100}
Marc~J. Dupuis and Andrew Williams. 2019.
\newblock The spread of disinformation on the {W}eb: An examination of memes on
  social networking.
\newblock In \emph{Proceedings of the 2019 IEEE SmartWorld, Ubiquitous
  Intelligence Computing, Advanced Trusted Computing, Scalable Computing
  Communications, Cloud Big Data Computing, Internet of People and Smart City
  Innovation}, SmartWorld/SCALCOM/UIC/ATC/CBDCom/IOP/SCI~'19, pages 1412--1418.

\bibitem[{Gundapu and Mamidi(2020)}]{gundapu-mamidi-2020-gundapusunil}
Sunil Gundapu and Radhika Mamidi. 2020.
\newblock Gundapusunil at {S}em{E}val-2020 task 8: {M}ultimodal {M}emotion
  {A}nalysis.
\newblock In \emph{Proceedings of the Fourteenth Workshop on Semantic
  Evaluation}, SemEval~'20, pages 1112--1119, Barcelona, Spain.

\bibitem[{Guo et~al.(2020)Guo, Huang, Dong, and Xu}]{guo-etal-2020-guoym}
Yingmei Guo, Jinfa Huang, Yanlong Dong, and Mingxing Xu. 2020.
\newblock Guoym at {S}em{E}val-2020 task 8: Ensemble-based classification of
  visuo-lingual metaphor in memes.
\newblock In \emph{Proceedings of the Fourteenth Workshop on Semantic
  Evaluation}, SemEval~'20, pages 1120--1125, Barcelona, Spain.

\bibitem[{Hawkes(1971)}]{Hawkes1971SpectraOS}
Alan Hawkes. 1971.
\newblock Spectra of some self-exciting and mutually exciting point processes.
\newblock \emph{Biometrika}, 58:83--90.

\bibitem[{He et~al.(2016)He, Zhang, Ren, and Sun}]{he2016deep}
Kaiming He, Xiangyu Zhang, Shaoqing Ren, and Jian Sun. 2016.
\newblock Deep {r}esidual {l}earning for {i}mage {r}ecognition.
\newblock In \emph{Proceedings of the IEEE Conference on Computer Vision and
  Pattern Recognition}, CVPR~'16, pages 770--778, Las Vegas, NV, USA.

\bibitem[{Hu and Flaxman(2018)}]{Hu_2018}
Anthony Hu and Seth Flaxman. 2018.
\newblock Multimodal sentiment analysis to explore the structure of emotions.
\newblock In \emph{Proceedings of the 24th ACM SIGKDD International Conference
  on Knowledge Discovery \& Data Mining}, ACMKDD~'18, pages 350--358, London,
  UK.

\bibitem[{Huang et~al.(2017)Huang, Liu, Van Der~Maaten, and
  Weinberger}]{huang2017densely}
Gao Huang, Zhuang Liu, Laurens Van Der~Maaten, and Kilian~Q. Weinberger. 2017.
\newblock Densely connected convolutional networks.
\newblock In \emph{Proceedings of the 2017 IEEE {C}onference on {C}omputer
  {V}ision and {P}attern {R}ecognition}, CVPR~'17, pages 2261--2269, Honolulu,
  HI, USA.

\bibitem[{Karkkainen and Joo(2021)}]{karkkainen2019fairface}
Kimmo Karkkainen and Jungseock Joo. 2021.
\newblock Fairface: Face attribute dataset for balanced race, gender, and age
  for bias measurement and mitigation.
\newblock In \emph{Proceedings of the IEEE/CVF Winter Conference on
  Applications of Computer Vision}, WACV~'21, pages 1548--1558.

\bibitem[{Kiela et~al.(2019)Kiela, Bhooshan, Firooz, and
  Testuggine}]{kiela2019supervised}
Douwe Kiela, Suvrat Bhooshan, Hamed Firooz, and Davide Testuggine. 2019.
\newblock Supervised multimodal bitransformers for classifying images and text.
\newblock \emph{arXiv 1909.02950}.

\bibitem[{Kiela et~al.(2020)Kiela, Firooz, Mohan, Goswami, Singh, Ringshia, and
  Testuggine}]{kiela2020hateful}
Douwe Kiela, Hamed Firooz, Aravind Mohan, Vedanuj Goswami, Amanpreet Singh,
  Pratik Ringshia, and Davide Testuggine. 2020.
\newblock The hateful memes challenge: Detecting hate speech in multimodal
  memes.
\newblock In \emph{Advances in Neural Information Processing Systems},
  volume~33 of \emph{NeurIPS~'20}, pages 2611--2624, Vancouver, Canada.

\bibitem[{Kingma and Ba(2014)}]{kingma2014adam}
Diederik~P Kingma and Jimmy Ba. 2014.
\newblock Adam: A method for stochastic optimization.
\newblock In \emph{Proceedings of the 3rd International Conference on Learning
  Representations}, ICLR~'14, San Diego, CA, USA.

\bibitem[{Leskovec et~al.(2009)Leskovec, Backstrom, and
  Kleinberg}]{10.1145/1557019.1557077}
Jure Leskovec, Lars Backstrom, and Jon Kleinberg. 2009.
\newblock Meme-tracking and the dynamics of the news cycle.
\newblock In \emph{Proceedings of the 15th ACM SIGKDD International Conference
  on Knowledge Discovery and Data Mining}, KDD '09, page 497–506, New York,
  USA.

\bibitem[{Li et~al.(2019)Li, Yatskar, Yin, Hsieh, and Chang}]{li2019visualbert}
Liunian~Harold Li, Mark Yatskar, Da~Yin, Cho-Jui Hsieh, and Kai-Wei Chang.
  2019.
\newblock {VisualBERT}: A simple and performant baseline for vision and
  language.
\newblock \emph{arxiv 1908.03557}.

\bibitem[{Li et~al.(2020)Li, Yin, Li, Zhang, Hu, Zhang, Wang, Hu, Dong, Wei,
  Choi, and Gao}]{li2020oscar}
Xiujun Li, Xi~Yin, Chunyuan Li, Pengchuan Zhang, Xiaowei Hu, Lei Zhang, Lijuan
  Wang, Houdong Hu, Li~Dong, Furu Wei, Yejin Choi, and Jianfeng Gao. 2020.
\newblock Oscar: Object-semantics aligned pre-training for vision-language
  tasks.
\newblock In \emph{Proceedings of the 16th European Computer Vision
  Conference}, ECCV~'20, pages 121--137, Glasgow, UK.

\bibitem[{Lin et~al.(2014)Lin, Maire, Belongie, Hays, Perona, Ramanan,
  Doll{\'a}r, and Zitnick}]{lin2014microsoft}
Tsung-Yi Lin, Michael Maire, Serge Belongie, James Hays, Pietro Perona, Deva
  Ramanan, Piotr Doll{\'a}r, and C~Lawrence Zitnick. 2014.
\newblock Microsoft {COCO}: {C}ommon {O}bjects in {C}ontext.
\newblock In \emph{Proceedings of the European Conference on Computer Vision},
  ECCV~'14, pages 740--755, Zurich, Switzerland.

\bibitem[{Ling et~al.(2021)Ling, AbuHilal, Blackburn, Cristofaro, Zannettou,
  and Stringhini}]{ling2021dissecting}
Chen Ling, Ihab AbuHilal, Jeremy Blackburn, Emiliano~De Cristofaro, Savvas
  Zannettou, and Gianluca Stringhini. 2021.
\newblock Dissecting the meme magic: Understanding indicators of virality in
  image memes.
\newblock In \emph{Proceedings of the 24th ACM Conference on Computer-Supported
  Cooperative Work and Social Computing}, CSCW~'21.

\bibitem[{Lippe et~al.(2020)Lippe, Holla, Chandra, Rajamanickam, Antoniou,
  Shutova, and Yannakoudakis}]{lippe2020multimodal}
Phillip Lippe, Nithin Holla, Shantanu Chandra, Santhosh Rajamanickam, Georgios
  Antoniou, Ekaterina Shutova, and Helen Yannakoudakis. 2020.
\newblock A multimodal framework for the detection of hateful memes.
\newblock \emph{arXiv 2012.12871}.

\bibitem[{Liu et~al.(2020)Liu, Osei-Brefo, Chen, and Liang}]{liu-etal-2020-uor}
Zehao Liu, Emmanuel Osei-Brefo, Siyuan Chen, and Huizhi Liang. 2020.
\newblock {U}o{R} at {S}em{E}val-2020 task 8: {G}aussian mixture modelling
  ({GMM}) based sampling approach for multi-modal memotion analysis.
\newblock In \emph{Proceedings of the Fourteenth Workshop on Semantic
  Evaluation}, SemEval~'20, pages 1201--1207, Barcelona, Spain.

\bibitem[{Lu et~al.(2019)Lu, Batra, Parikh, and Lee}]{lu2019vilbert}
Jiasen Lu, Dhruv Batra, Devi Parikh, and Stefan Lee. 2019.
\newblock {ViLBERT}: Pretraining task-agnostic visiolinguistic representations
  for vision-and-language tasks.
\newblock In \emph{Proceedings of the Conference on Neural Information
  Processing Systems}, NeurIPS~'19, pages 13--23, Vancouver, Canada.

\bibitem[{Mittos et~al.(2020)Mittos, Zannettou, Blackburn, and
  Cristofaro}]{mittos2019and}
Alexandros Mittos, Savvas Zannettou, Jeremy Blackburn, and Emiliano~De
  Cristofaro. 2020.
\newblock ``{A}nd we will fight for our race!'' {A} measurement study of
  genetic testing conversations on {R}eddit and 4chan.
\newblock In \emph{Proceedings of the {F}ourteenth {I}nternational {AAAI}
  {C}onference on {W}eb and {S}ocial {M}edia}, ICWSM~'20, pages 452--463,
  Atlanta, GA, USA.

\bibitem[{Morishita et~al.(2020)Morishita, Morio, Horiguchi, Ozaki, and
  Miyoshi}]{morishita-etal-2020-hitachi-semeval-2020}
Terufumi Morishita, Gaku Morio, Shota Horiguchi, Hiroaki Ozaki, and Toshinori
  Miyoshi. 2020.
\newblock Hitachi at {S}em{E}val-2020 task 8: Simple but effective modality
  ensemble for meme emotion recognition.
\newblock In \emph{Proceedings of the Fourteenth Workshop on Semantic
  Evaluation}, SemEval~'20, pages 1126--1134, Barcelona, Spain.

\bibitem[{Muennighoff(2020)}]{muennighoff2020vilio}
Niklas Muennighoff. 2020.
\newblock Vilio: {S}tate-of-the-art {V}isio-{L}inguistic {M}odels applied to
  {H}ateful {M}emes.
\newblock \emph{arxiv 2012.07788}.

\bibitem[{De~la Pe{\~n}a~Sarrac{\'e}n et~al.(2020)De~la Pe{\~n}a~Sarrac{\'e}n,
  Rosso, and Giachanou}]{de-la-pena-sarracen-etal-2020-prhlt}
Gretel~Liz De~la Pe{\~n}a~Sarrac{\'e}n, Paolo Rosso, and Anastasia Giachanou.
  2020.
\newblock {PRHLT}-{UPV} at {S}em{E}val-2020 task 8: Study of multimodal
  techniques for memes analysis.
\newblock In \emph{Proceedings of the Fourteenth Workshop on Semantic
  Evaluation}, SemEval~'20, pages 908--915, Barcelona, Spain.

\bibitem[{Pramanick et~al.(2021)Pramanick, Akhtar, and
  Chakraborty}]{pramanick2021exercise}
Shraman Pramanick, Md~Shad Akhtar, and Tanmoy Chakraborty. 2021.
\newblock Exercise? {I} thought you said `extra fries': Leveraging sentence
  demarcations and multi-hop attention for meme affect analysis.
\newblock \emph{proceedings of the Fifteenth International AAAI Conference on
  Web and Social Media}.

\bibitem[{Reis et~al.(2020)Reis, Melo, Garimella, Almeida, Eckles, and
  Benevenuto}]{reis2020dataset}
Julio C.~S. Reis, Philipe Melo, Kiran Garimella, Jussara~M. Almeida, Dean
  Eckles, and Fabrício Benevenuto. 2020.
\newblock A dataset of fact-checked images shared on {WhatsApp} during the
  {B}razilian and {I}ndian elections.
\newblock In \emph{Proceedings of the International AAAI Conference on Web and
  Social Media}, ICWSM~'20, pages 903--908.

\bibitem[{Ribeiro et~al.(2016)Ribeiro, Singh, and Guestrin}]{ribeiro2016should}
Marco~Tulio Ribeiro, Sameer Singh, and Carlos Guestrin. 2016.
\newblock ``{W}hy should {I} trust you?'' {E}xplaining the predictions of any
  classifier.
\newblock In \emph{Proceedings of the 22nd ACM SIGKDD International Conference
  on Knowledge Discovery and Data Mining}, KDD~'16, pages 1135--1144, San
  Francisco, CA, USA.

\bibitem[{Sabat et~al.(2019)Sabat, Canton{-}Ferrer, and
  Gir{\'{o}}{-}i{-}Nieto}]{DBLP:journals/corr/abs-1910-02334}
Benet~Oriol Sabat, Cristian Canton{-}Ferrer, and Xavier Gir{\'{o}}{-}i{-}Nieto.
  2019.
\newblock Hate speech in pixels: Detection of offensive memes towards automatic
  moderation.
\newblock \emph{arXiv 1910.02334}.

\bibitem[{Sandulescu(2020)}]{sandulescu2020detecting}
Vlad Sandulescu. 2020.
\newblock Detecting hateful memes using a multimodal deep ensemble.
\newblock \emph{arXiv 2012.13235}.

\bibitem[{Sharma et~al.(2020{\natexlab{a}})Sharma, Bhageria, Scott, PYKL, Das,
  Chakraborty, Pulabaigari, and Gamb{\"a}ck}]{sharma-etal-2020-semeval}
Chhavi Sharma, Deepesh Bhageria, William Scott, Srinivas PYKL, Amitava Das,
  Tanmoy Chakraborty, Viswanath Pulabaigari, and Bj{\"o}rn Gamb{\"a}ck.
  2020{\natexlab{a}}.
\newblock {S}em{E}val-2020 task 8: Memotion analysis- the visuo-lingual
  metaphor!
\newblock In \emph{Proceedings of the Fourteenth Workshop on Semantic
  Evaluation}, SemEval~'20, pages 759--773, Barcelona, Spain.

\bibitem[{Sharma et~al.(2020{\natexlab{b}})Sharma, Kandasamy, and
  Vasantha}]{sharma-etal-2020-memebusters}
Mayukh Sharma, Ilanthenral Kandasamy, and W.b. Vasantha. 2020{\natexlab{b}}.
\newblock Memebusters at {S}em{E}val-2020 task 8: Feature fusion model for
  sentiment analysis on memes using transfer learning.
\newblock In \emph{Proceedings of the Fourteenth Workshop on Semantic
  Evaluation}, SemEval~'20, pages 1163--1171, Barcelona, Spain.

\bibitem[{Sharma et~al.(2018)Sharma, Ding, Goodman, and
  Soricut}]{sharma2018conceptual}
Piyush Sharma, Nan Ding, Sebastian Goodman, and Radu Soricut. 2018.
\newblock Conceptual captions: A cleaned, hypernymed, image alt-text dataset
  for automatic image captioning.
\newblock In \emph{Proceedings of the 56th Annual Meeting of the Association
  for Computational Linguistics}, ACL~'18, pages 2556--2565, Melbourne,
  Australia.

\bibitem[{Simonyan and Zisserman(2015)}]{simonyan2014very}
Karen Simonyan and Andrew Zisserman. 2015.
\newblock Very deep convolutional networks for large-scale image recognition.
\newblock In \emph{Proceedings of the 3rd International Conference on Learning
  Representations}, ICLR~'15, San Diego, CA, USA.

\bibitem[{Singh et~al.(2020)Singh, Goswami, and Parikh}]{singh2020we}
Amanpreet Singh, Vedanuj Goswami, and Devi Parikh. 2020.
\newblock Are we pretraining it right? {D}igging deeper into visio-linguistic
  pretraining.
\newblock \emph{arXiv 2004.08744}.

\bibitem[{Strubell et~al.(2019)Strubell, Ganesh, and
  McCallum}]{strubell-etal-2019-energy}
Emma Strubell, Ananya Ganesh, and Andrew McCallum. 2019.
\newblock Energy and policy considerations for deep learning in {NLP}.
\newblock In \emph{Proceedings of the 57th Annual Meeting of the Association
  for Computational Linguistics}, ACL~'19, pages 3645--3650, Florence, Italy.

\bibitem[{Su et~al.(2020)Su, Zhu, Cao, Li, Lu, Wei, and Dai}]{su2020vlbert}
Weijie Su, Xizhou Zhu, Yue Cao, Bin Li, Lewei Lu, Furu Wei, and Jifeng Dai.
  2020.
\newblock {VL-BERT:} pre-training of generic visual-linguistic representations.
\newblock In \emph{Proceedings of the 8th International Conference on Learning
  Representations}, ICLR~'20, Addis Ababa, Ethiopia.

\bibitem[{Suryawanshi et~al.(2020)Suryawanshi, Chakravarthi, Arcan, and
  Buitelaar}]{suryawanshi-etal-2020-multimodal}
Shardul Suryawanshi, Bharathi~Raja Chakravarthi, Mihael Arcan, and Paul
  Buitelaar. 2020.
\newblock Multimodal meme dataset ({M}ulti{OFF}) for identifying offensive
  content in image and text.
\newblock In \emph{Proceedings of the Second Workshop on Trolling, Aggression
  and Cyberbullying}, LREC-TRAC~'20, pages 32--41, Marseille, France.

\bibitem[{Tan and Bansal(2019)}]{tan2019lxmert}
Hao Tan and Mohit Bansal. 2019.
\newblock {LXMERT}: Learning cross-modality encoder representations from
  transformers.
\newblock In \emph{{P}roceedings of the 2019 {C}onference on {E}mpirical
  {M}ethods in {N}atural {L}anguage {P}rocessing and the 9th {I}nternational
  {J}oint {C}onference on {N}atural {L}anguage {P}rocessing},
  EMNLP-IJCNLP~’17, pages 5100--5111, Hong Kong, China.

\bibitem[{{Thang Duong} et~al.(2017){Thang Duong}, {Lebret}, and
  {Aberer}}]{duong2017multimodal}
Chi {Thang Duong}, Remi {Lebret}, and Karl {Aberer}. 2017.
\newblock Multimodal classification for analysing social media.
\newblock \emph{arXiv 1708.02099}.

\bibitem[{Vaswani et~al.(2017)Vaswani, Shazeer, Parmar, Uszkoreit, Jones,
  Gomez, Kaiser, and Polosukhin}]{vaswani2017attention}
Ashish Vaswani, Noam Shazeer, Niki Parmar, Jakob Uszkoreit, Llion Jones,
  Aidan~N Gomez, Lukasz Kaiser, and Illia Polosukhin. 2017.
\newblock Attention is all you need.
\newblock In \emph{Proceedings of the Annual Conference on Neural Information
  Processing Systems}, NeurIPS~'17, pages 5998--6008, Long Beach, CA, USA.

\bibitem[{Wang et~al.(2020)Wang, Tahmasbi, Blackburn, Bradlyn, Cristofaro,
  Magerman, Zannettou, and Stringhini}]{wang2020understanding}
Yuping Wang, Fatemeh Tahmasbi, Jeremy Blackburn, Barry Bradlyn, Emiliano~De
  Cristofaro, David Magerman, Savvas Zannettou, and Gianluca Stringhini. 2020.
\newblock Understanding the use of fauxtography on social media.
\newblock \emph{arXiv 2009.11792}.

\bibitem[{Xie et~al.(2017)Xie, Girshick, Doll{\'a}r, Tu, and
  He}]{xie2017aggregated}
Saining Xie, Ross Girshick, Piotr Doll{\'a}r, Zhuowen Tu, and Kaiming He. 2017.
\newblock Aggregated residual transformations for deep neural networks.
\newblock In \emph{Proceedings of the {IEEE} Conference on Computer Vision and
  Pattern Recognition}, CVPR~’17, pages 1492--1500, Honolulu, HI, USA.

\bibitem[{Yu et~al.(2021)Yu, Tang, Yin, Sun, Tian, Wu, and
  Wang}]{yu2020ernievil}
Fei Yu, Jiji Tang, Weichong Yin, Yu~Sun, Hao Tian, Hua Wu, and Haifeng Wang.
  2021.
\newblock {ERNIE-ViL}: Knowledge enhanced vision-language representations
  through scene graph.
\newblock In \emph{Proceedings of the Thirty-Fifth {AAAI} Conference on
  Artificial Intelligence}, AAAI~'21, pages 3208--3216.

\bibitem[{Yuan et~al.(2020)Yuan, Wang, and Zhang}]{yuan-etal-2020-ynu}
Li~Yuan, Jin Wang, and Xuejie Zhang. 2020.
\newblock {YNU}-{HPCC} at {S}em{E}val-2020 task 8: Using a parallel-channel
  model for memotion analysis.
\newblock In \emph{Proceedings of the Fourteenth Workshop on Semantic
  Evaluation}, SemEval~'20, pages 916--921, Barcelona, Spain.

\bibitem[{Zannettou et~al.(2020{\natexlab{a}})Zannettou, Caulfield, Bradlyn,
  Cristofaro, Stringhini, and Blackburn}]{zannettou2019characterizing}
Savvas Zannettou, Tristan Caulfield, Barry Bradlyn, Emiliano~De Cristofaro,
  Gianluca Stringhini, and Jeremy Blackburn. 2020{\natexlab{a}}.
\newblock Characterizing the use of images in state-sponsored information
  warfare operations by {R}ussian trolls on {T}witter.
\newblock In \emph{Proceedings of the {F}ourteenth {I}nternational {AAAI}
  {C}onference on {W}eb and {S}ocial {M}edia}, ICWSM~’20, pages 774--785,
  Atlanta, GA, USA.

\bibitem[{Zannettou et~al.(2020{\natexlab{b}})Zannettou, Finkelstein, Bradlyn,
  and Blackburn}]{zannettou2019quantitative}
Savvas Zannettou, Joel Finkelstein, Barry Bradlyn, and Jeremy Blackburn.
  2020{\natexlab{b}}.
\newblock A quantitative approach to understanding online antisemitism.
\newblock In \emph{Proceedings of the {F}ourteenth {I}nternational {AAAI}
  {C}onference on {W}eb and {S}ocial {M}edia}, ICWSM~'20, pages 786--797,
  Atlanta, GA, USA.

\bibitem[{{Zhong}(2020)}]{zhong2020classification}
Xiayu {Zhong}. 2020.
\newblock Classification of multimodal hate speech -- the winning solution of
  hateful memes challenge.
\newblock \emph{arXiv 2012.01002}.

\bibitem[{Zhou et~al.(2020)Zhou, Palangi, Zhang, Hu, Corso, and
  Gao}]{zhou2019unified}
Luowei Zhou, Hamid Palangi, Lei Zhang, Houdong Hu, Jason Corso, and Jianfeng
  Gao. 2020.
\newblock Unified vision-language pre-training for image captioning and {VQA}.
\newblock \emph{Proceedings of the AAAI Conference on Artificial Intelligence},
  pages 13041--13049.

\bibitem[{Zhou and Chen(2020)}]{zhou2020multimodal}
Yi~Zhou and Zhenhao Chen. 2020.
\newblock Multimodal learning for hateful memes detection.
\newblock \emph{arXiv 2011.12870}.

\bibitem[{Zhu(2020)}]{zhu2020enhance}
Ron Zhu. 2020.
\newblock Enhance multimodal transformer with external label and in-domain
  pretrain: Hateful meme challenge winning solution.
\newblock \emph{arXiv 2012.08290}.

\end{thebibliography}

\clearpage
\newpage
\appendix

\counterwithin{figure}{section}
\numberwithin{table}{section}

\section{Implementation Details and Hyper-Parameter Values} \label{sec:hyperparameters}

We trained all the models using the Pytorch framework on an NVIDIA Tesla T4 GPU with 16 GB of dedicated memory and with CUDA-10 and cuDNN-11 installed. For the unimodal models, we imported all the pre-trained weights from the TORCHVISION.MODELS\footnote{http://pytorch.org/docs/stable/torchvision/models.html} subpackage of PyTorch. We initialized the non pre-trained weights randomly with a zero-mean Gaussian distribution with a standard deviation of 0.02.
To minimize the impact of the label imbalance in the loss calculation, we assigned larger weights to the minority class. We trained our models using the Adam optimizer \cite{kingma2014adam} and the negative log-likelihood loss as the objective function. Table~\ref{tab:hyperparameters} gives the values of all hyper-parameters we used for training.

We trained the models end-to-end for the two classification tasks, i.e.,~the memes that were classified as \textit{Very Harmful} or \textit{Partially Harmful} in the first classification stage were sent to the second stage for target identification. 

\begin{table*}[ht!]
\centering
\resizebox{\textwidth}{!}
{
\begin{tabular}{c | l | c | c | c | c | c | c }
\hline
& \multirow{2}{*}{\bf Models} & \multicolumn{6}{c}{\bf Hyper-parameters} \\ \cline{3-8}
& &  \bf Batch Size & \bf Epochs & \bf Learning Rate & \bf Image Encoder & \bf Text Encoder & \bf \#Parameters \\ \hline

\multirow{5}{*}{\begin{turn}{-90} \centering \bf Unimodal \end{turn}} & TextBERT & 16 & 100 & 0.001 & - & Bert-base-uncased & 110,683,414\\
& VGG19 & 64 & 200 & 0.01 & VGG19 & - & 138,357,544\\
& DenseNet-161 & 32 & 200 & 0.01 & DenseNet-161 & - & 28,681,538\\
& ResNet-152 & 32 & 300 & 0.01 & ResNet-152 & - & 60,192,808\\
& ResNeXt-101 & 32 & 300 & 0.01 & ResNeXt-101 & - & 83,455,272\\
\hline 
\multirow{5}{*}{\begin{turn}{-90} \centering \bf Multimodal \end{turn}} & Late Fusion & 16 & 200 & 0.0001 & ResNet-152 & Bert-base-uncased & 170,983,752 \\
& Concat BERT & 16 & 200 & 0.001 & ResNet-152 & Bert-base-uncased & 170,982,214\\
& MMBT & 16 & 200 & 0.001 & ResNet-152 & Bert-base-uncased & 169,808,726\\
& ViLBERT CC & 16 & 100 & 0.001 & Faster RCNN & Bert-base-uncased & 112,044,290\\
& V-BERT COCO & 16 & 100 & 0.001 & Faster RCNN & Bert-base-uncased & 247,782,404\\
\hline
\end{tabular}
}
\caption{The values of the hyper-parameters of all our models.}
\label{tab:hyperparameters}
\end{table*}
    
\section{Annotation Guidelines} \label{sec:annotation}

\subsection{What do we mean by \textit{harmful} memes?}
The entrenched meaning of harmful memes is targeted towards a social entity (e.g.,~an individual, an organization, a community, etc.), likely to cause calumny/vilification/defamation depending on their background (bias, social background, educational background, etc.). The \emph{harm} caused by a meme can be in the form of mental abuse, psycho-physiological injury, proprietary damage, emotional disturbance, compensated public image. A harmful meme typically attacks celebrities or well-known organizations, with the intent to expose their professional demeanor.

\bigskip
\noindent\textbf{Characteristics of \textit{harmful} memes:}

\begin{itemize}[leftmargin=*]
    \item Harmful memes may or may not be offensive, hateful, or biased in nature.
    \item Harmful memes expose vices, allegations, and other negative aspects of an entity based on verified or unfounded claims or mocks. 
    \item Harmful memes leave an open-ended connotation to the word \emph{community}, including \emph{antisocial} communities such as terrorist groups.
    \item The harmful content in harmful memes is often implicit and might require critical judgment to establish its potential to do harm.
    \item Harmful memes can be classified at multiple levels, based on the intensity of the harm they could cause, e.g.,~\textit{very harmful} or \textit{partially harmful}. 
    \item One harmful meme can target multiple individuals, organizations, and/or communities at the same time. In that case, we asked the annotators to go with the best personal judgment. 
    \item Harm can be expressed in the form of sarcasm and/or political satire. Sarcasm is praise that is actually an insult; sarcasm generally involves malice, the desire to put someone down. On the other hand, satire is the ironical exposure of the vices or the follies of an individual, a group, an institution, an idea, the society, etc., usually with the aim to correcting it.

\end{itemize}

\begin{figure}[ht!]
\centering
\subfloat[{Non-English (Hindi) meme. \imgsrc{https://i.imgur.com/reikEuL.jpeg}{} \imgur{}} \label{fig:meme_rej:exm:1}]{
\includegraphics[width=0.232\textwidth, height=0.225\textwidth]{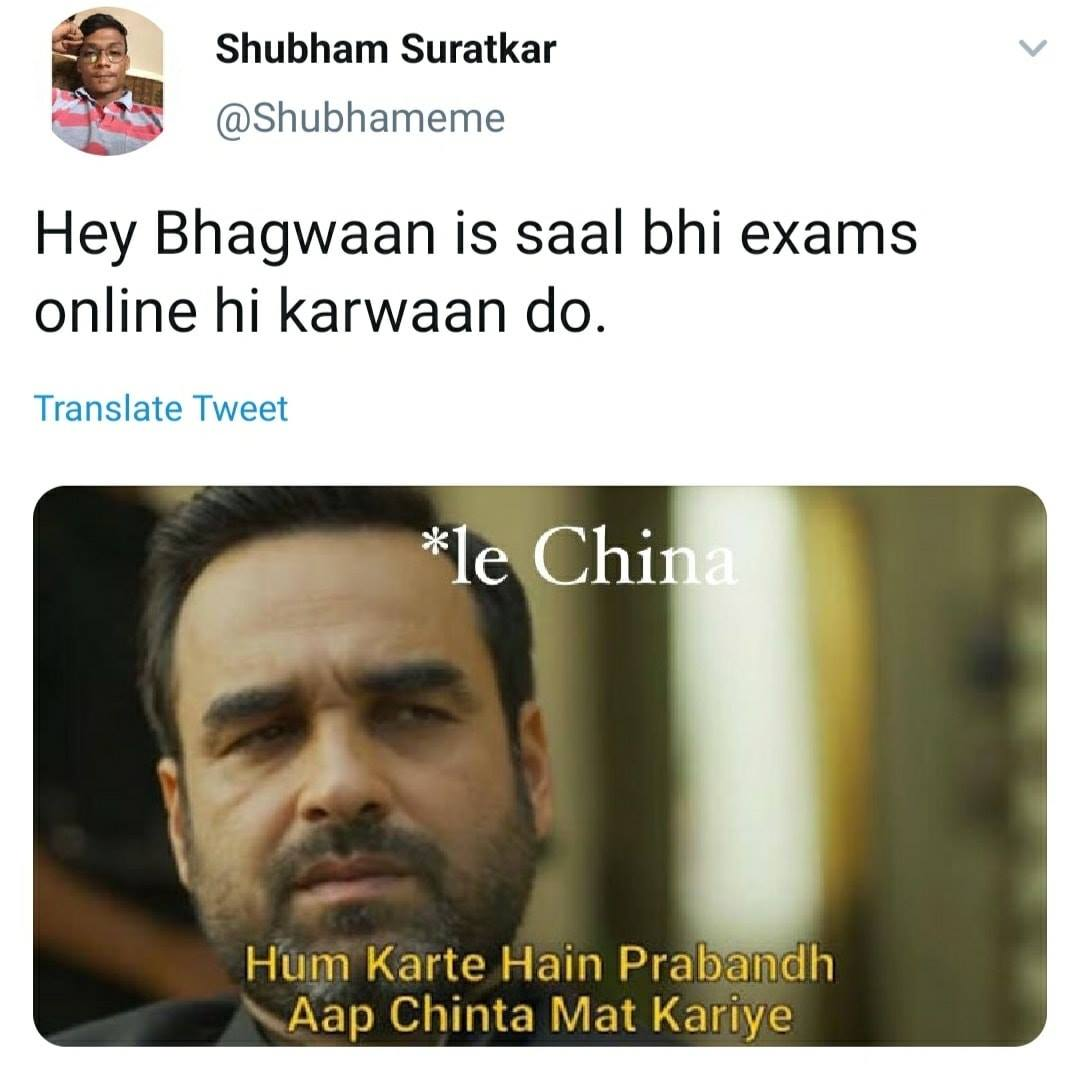}}\hspace{0.1em}
\subfloat[{\centering Unreadable meme. \imgsrc{https://i.imgur.com/5RwUS24.jpeg}{} \ccsnd{}}\vspace{9pt} \label{fig:meme_rej:exm:2}]{
\includegraphics[width=0.228\textwidth, height=0.225\textwidth]{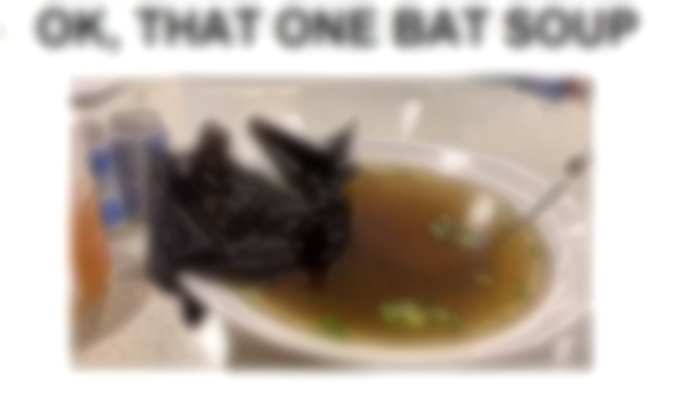}}\hspace{0.1em}
\subfloat[{\centering Meme with a cartoon. \imgsrc{https://search.creativecommons.org/photos/dd244e4f-d472-4e2b-8c69-24532095f10f}{} \imgur{}} \label{fig:meme_rej:exm:3}]{
\includegraphics[width=0.230\textwidth, height=0.225\textwidth]{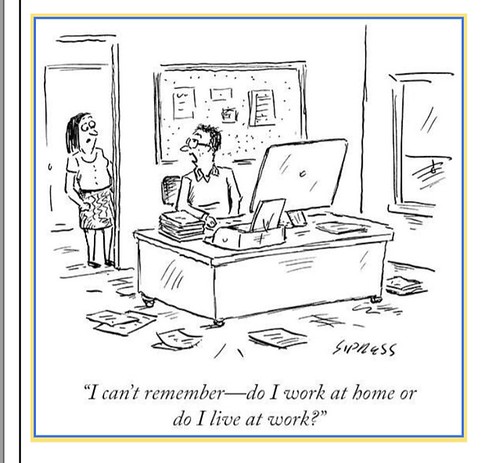}}\hspace{0.2em}
\subfloat[{\centering Meme without textual modality. \imgsrc{https://i.imgur.com/abgvmUo.jpg}{} \imgur{}}\vspace{9pt} \label{fig:meme_rej:exm:4a}]{
\includegraphics[width=0.228\textwidth, height=0.225\textwidth]{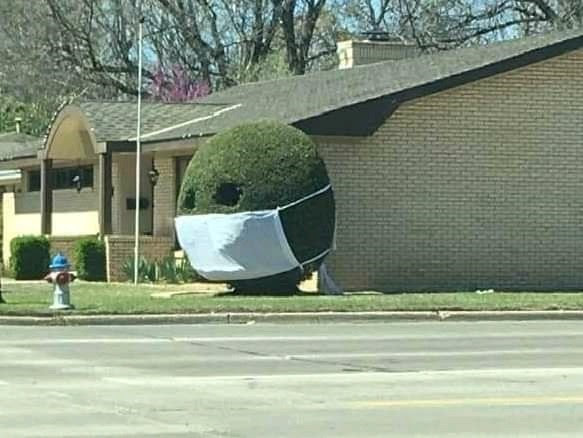}}\hspace{0.1em}
\subfloat[{\centering Meme without visual modality. \imgsrc{https://i.imgur.com/tFhGulA.jpg}{} \imgur{}} \label{fig:meme_rej:exm:4b}]{
\includegraphics[width=0.35\textwidth, height=0.225\textwidth]{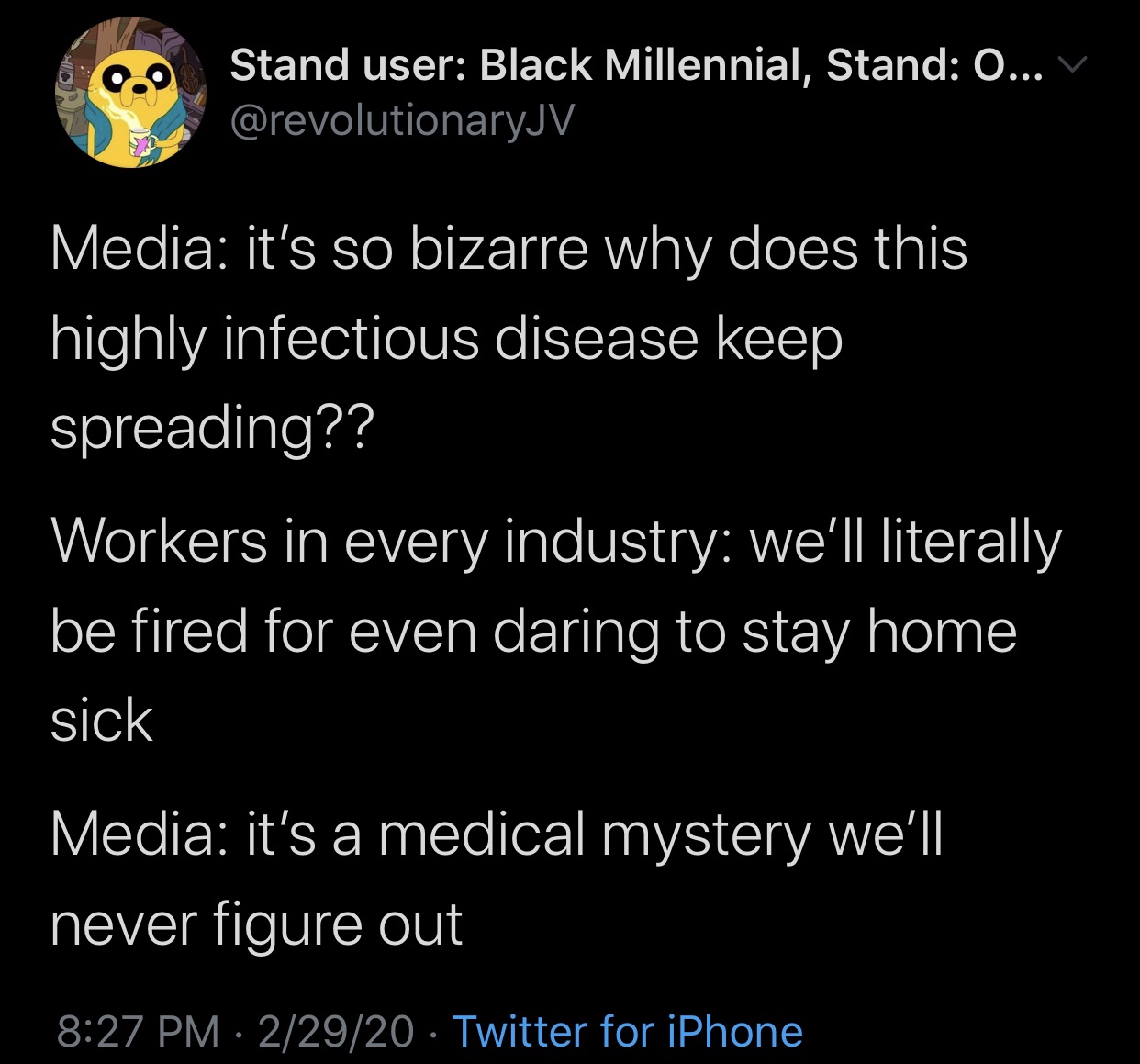}}\hspace{0.1em}
\caption{Examples of memes that we rejected during the process of data collection and annotation.}
\label{fig:meme:filtered_example}
\end{figure}

\subsection{What is the difference between \textit{organization} and \textit{community}?}

An organization is a group of people with a particular purpose, such as a business or a government department. Examples include a company, an institution, or an association comprising one or more people with a particular purpose, e.g.,~a research organization, a political organization, etc.

On the other hand, a community is a social unit (a group of living things) with a commonality such as norms, religion, values, ideology customs, or identity. Communities may share a sense of place situated in a given geographical area (e.g.,~a country, a village, a town, or a neighborhood) or in the virtual space through communication platforms.

\subsection{When do we \textit{reject} a meme?} \label{sec:filtering}

We apply the following rejection criteria during the process of data collection and annotation:

\begin{enumerate}
    \item The meme's text is code-mixed or not in English.
    \item The meme's text is not readable. (e.g.,~blurry text, incomplete text, etc.)
    \item The meme is unimodal in nature, containing only textual or only visual content.
    \item The meme contains a cartoon.
\end{enumerate}

Figure \ref{fig:meme:filtered_example} shows some rejected memes.

\end{document}

\maketitle

\section{Implementation Details and Hyperparameters}

We train all the models using Pytorch \citep{paszke2019pytorch} framework on a NVIDIA Tesla T4 GPU, with 16 GB dedicated memory, with CUDA-10 and cuDNN-11 installed. For the unimodal  models, We import all the pre-trained weights from TORCHVISION.MODELS\footnote{https://pytorch.org/docs/stable/torchvision/models.html} subpackage of PyTorch framework. The non pre-trained weights are randomly initialized with a zero-mean Gaussian distribution with standard deviation 0.02. From the dataset statistics table, we can observe label imbalance problem for both harmfulness intensity ([\textit{Very Harmful}, \textit{Partially Harmful}] vs. \textit{Harmless}) and target ([\textit{Individual}, \textit{Organization}, \textit{Community}] vs. \textit{Entire Society}) classification tasks. We train the  models end-to-end for the two classification tasks, i.e.,  memes which are classified as \textit{Very Harmful} or \textit{Partially Harmful} in the first classification stage, are sent into the second stage for target identification. 
To minimize the effect of label imbalance in loss calculation, we assign larger weights for minority class. We train our models using Adam \cite{kingma2014adam} optimizer and negative log-likelihood (NLL) loss as the objective function. In Table \ref{tab:hyperparameters}, we furnish the details of hyper-parameters used for the training.

\begin{table}[h]
\centering
\resizebox{0.99\columnwidth}{!}
{
\begin{tabular}{C | l | c | c | c | c | c | c }
& \multirow{2}{*}{\bf Models} & \multicolumn{6}{c}{\bf Hyperparameters} \\ \cline{3-8}
& &  \bf Batch-size & \bf Epochs & \bf Learning Rate & \bf Image Encoder & \bf Text Encoder & \bf \#Parameters \\ \hline

\multirow{5}{*}{\begin{turn}{-90} \centering \bf Unimodal \end{turn}} & TextBERT & 16 & 100 & 0.001 & - & Bert-base-uncased & 110,683,414\\
& VGG19 & 64 & 200 & 0.01 & VGG19 & - & 138,357,544\\
& DenseNet-161 & 32 & 200 & 0.01 & DenseNet-161 & - & 28,681,538\\
& ResNet-152 & 32 & 300 & 0.01 & ResNet-152 & - & 60,192,808\\
& ResNeXt-101 & 32 & 300 & 0.01 & ResNeXt-101 & - & 83,455,272\\
\hline 
\multirow{5}{*}{\begin{turn}{-90} \centering \bf Multimodal \end{turn}} & Late Fusion & 16 & 200 & 0.0001 & ResNet-152 & Bert-base-uncased & 170,983,752 \\
& Concat BERT & 16 & 200 & 0.001 & ResNet-152 & Bert-base-uncased & 170,982,214\\
& MMBT & 16 & 200 & 0.001 & ResNet-152 & Bert-base-uncased & 169,808,726\\
& ViLBERT CC & 16 & 100 & 0.001 & Faster RCNN & Bert-base-uncased & 112,044,290\\
& V-BERT COCO & 16 & 100 & 0.001 & Faster RCNN & Bert-base-uncased & 247,782,404\\

\hline
\end{tabular}}

\caption{Hyperparameters of different models.}

\label{tab:hyperparameters}
\vspace{-2mm}
\end{table}
    
\section{Annotation Guidelines}

\subsection{What do we mean by \textit{harmful} memes?}
The entrenched meaning of harmful memes is targeted towards a social entity (e.g., an individual, an organization, a community, etc.), likely to cause calumny/ vilification/ defamation depending on their background (bias, social background, educational background, etc.). The ‘harm’ caused by a meme can be in the form of mental abuse, psycho-physiological injury, proprietary damage, emotional disturbance, compensated public image. A harmful meme typically attacks celebrities or well-known organizations intending to expose their professional demeanor.

\bigskip
\textbf{Characteristics of \textit{harmful} memes:}

\begin{itemize}
    \item Harmful memes may or may not be offensive, hateful or biased in nature.
    \item Harmful memes point out vices, allegations, and other negative aspects of an entity based on verified or unfounded claims or mocks. 
    \item Harmful memes leave an open-ended connotation to the word ‘community’, including ‘antisocial’ communities such as terrorist groups.
    \item The harmful content in harmful memes is often implicit and might require critical judgment to establish the potency it can cause.
    \item Harmful memes can be classified on multiple levels, based on the intensity of the harm caused, e.g., \textit{very harmful}, \textit{partially harmful}. 
    \item One harmful meme can target multiple individual, organizations, communities at the same time. In that case, we asked the annotators to go with the best personal judgment. 
    \item Harm can be expressed in form of sarcasm and/or political satire. Sarcasm is praise which is really an insult; sarcasm generally involves malice, the desire to put someone down. On the other hand, satire is the ironical exposure of the vices or follies of an individual, a group, an institution, an idea, a society, etc., usually with a view to correcting it.

\end{itemize}

\subsection{What is the difference between \textit{organization} and \textit{community}?}

An organization is a group of people with a particular purpose, such as a business or government department. In other words, an organization is a company, an institution, or an association $–$ comprising one or more people and having a particular purpose, e.g., a research organization, a political organization.

On the other hand, a community is a social unit (a group of living things) with a commonality such as norms, religion, values, ideology customs, or identity. Communities may share a sense of place situated in a given geographical area (e.g. ,a country, village, town, or neighborhood) or in virtual space through communication platforms.

\begin{figure}[htp]
  \centering
  \scalebox{0.925}{
  \subfigure[Non-English meme.]{\includegraphics[width =0.50\columnwidth,height=0.40\columnwidth]{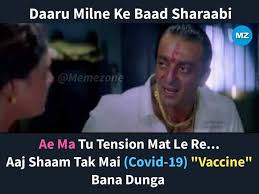}}}\hspace{0.5em}
  \scalebox{0.925}{
  \subfigure[Not-readable meme.]{\includegraphics[width =0.50\columnwidth,height=0.40\columnwidth]{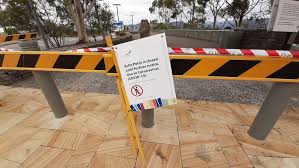}}}
  \scalebox{0.925}{
  \subfigure[Meme containing cartoon - often hard to interpret.]{\includegraphics[width =0.50\columnwidth,height=0.40\columnwidth]{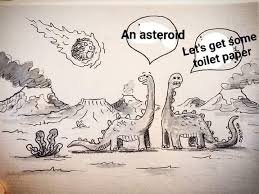}}}\hspace{0.5em}
  \scalebox{0.925}{
  \subfigure[Meme containing no textual modality.]{\includegraphics[width =0.50\columnwidth,height=0.40\columnwidth]{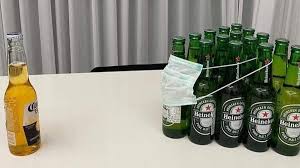}}}
  \subfigure[Meme containing no visual modality.]{\includegraphics[width =0.70\columnwidth,height=0.40\columnwidth]{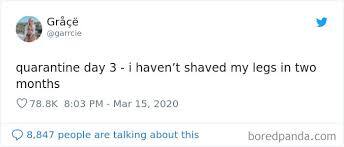}}
  \vspace{-2mm}
  \caption{Example of rejected memes during data collection and annotation process.}
  \label{figure:dataset_summary}
\end{figure}

\subsection{When do we \textit{reject} a meme?}

In order to assure a satisfactory quality of the dataset, we impose four well-defined rejection criterion during data collection and annotation process. The criterion are as follows:

\begin{enumerate}
    \item The meme text is in code-mixed or non-English language.
    \item The meme text is not readable. (e.g.,blurry text, incomplete text, etc.)
    \item The meme is unimodal in nature, containing only textual or visual content.
    \item The meme contains several cartoons. (we add this rejection criteria as cartoons are often very hard to be interpreted by AI systems)
\end{enumerate}

Figure \ref{figure:dataset_summary} presents a few rejected memes.

\bibliographystyle{acl_natbib}
\bibliography{acl2021}